\pdfoutput=1
\documentclass[11pt,dvipsnames, table]{article}

\usepackage[preprint]{acl}

\usepackage{times}
\usepackage{latexsym}

\usepackage{inconsolata}

\usepackage[T1]{fontenc}
\usepackage[utf8]{inputenc}
\usepackage{booktabs}
\usepackage{tabularx}
\usepackage{xcolor}
\usepackage{array}
\usepackage{amsmath}
\usepackage{bbm}
\usepackage{graphicx}
\usepackage{caption}
\usepackage{subcaption}
\usepackage{siunitx}
\usepackage{multirow}
\usepackage{url}
\usepackage{microtype}
\usepackage{makecell}
\usepackage{bbm}
\usepackage[english]{babel}
\usepackage{tikz}
\usepackage{listings}
\usepackage{amssymb}
\usepackage{fontawesome5} % for icons like \faBullseye

\usepackage[compact]{titlesec} % prevents weird gaps after seactino titles, helps with page fit

\usepackage[most,breakable]{tcolorbox}      % for text box

\lstdefinestyle{pyclean}{
  language=Python,
  basicstyle=\ttfamily\scriptsize,
  keywordstyle=\color{blue},
  commentstyle=\color{gray}\itshape,
  stringstyle=\color{green!50!black},
  showstringspaces=false,
  frame=lines,
  breaklines=true,
  tabsize=4,
  morekeywords={re, Counter}
}
\definecolor{expertiseColor}{HTML}{1d5f06}
\definecolor{robustnessColor}{HTML}{9900ff}
\definecolor{fidelityColor}{HTML}{af8713}

\newcommand{\baseline}{\emptyset}
\newcommand{\measure}{\textit{M}}

\newcommand{\task}{\mathit{T}}
\newcommand{\irrelevant}{\mathcal{I}_{\task}}

\newcommand{\personaSet}{\mathcal{P}}

\newcommand{\utility}{\textit{Adv}}
\newcommand{\worstcase}{\textit{Rob}}

\newcommand{\order}{\vec{O}}
\newcommand{\fidelity}{\textit{Fid}}
\newcommand{\inExpert}{\textit{exp}_\task}
\newcommand{\outExpert}{\textit{exp}_{\neg\task}}
\newcommand{\simExpert}{\textit{exp}_{{\sim}\task}}

\newcommand{\bExpert}{\textit{exp}_\textsc{Broad}}
\newcommand{\fExpert}{\textit{exp}_\textsc{Focused}}
\newcommand{\nExpert}{\textit{exp}_\textsc{Niche}}

\title{Principled Personas: Defining and Measuring the Intended Effects of Persona Prompting on Task Performance}

\author{Pedro Henrique Luz de Araujo\textsuperscript{{1,2}},
Paul Röttger\textsuperscript{{3}},
Dirk Hovy\textsuperscript{{3}}
\and
  Benjamin Roth\textsuperscript{{1,4}}
  \\
  \ \\
  \textsuperscript{1}University of Vienna, Faculty of Computer Science, Vienna, Austria
  \\
 \textsuperscript{2}Doctoral School Computer Science, University of Vienna, Vienna, Austria
  \\
  \textsuperscript{3}Bocconi University, Computing Sciences Department, Milan, Italy
  \\
 \textsuperscript{4}University of Vienna, Faculty of Philological and Cultural Studies, Vienna, Austria
  \\
  \texttt{\{pedro.henrique.luz.de.araujo, benjamin.roth\}@univie.ac.at} \\ 
}

\begin{document}
\maketitle
\begin{abstract}
  Expert persona prompting---assigning roles such as \emph{expert in math} to language models---is widely used for task improvement.
  However, prior work shows mixed results on its effectiveness, and does not consider when and why personas \textit{should} improve performance.
  We analyze the literature on persona prompting for task improvement and distill three desiderata: 1)~performance advantage of expert personas, 2)~robustness to irrelevant persona attributes, and 3)~fidelity to persona attributes.
  We then evaluate 9 state-of-the-art LLMs across 27 tasks with respect to these desiderata. 
  We find that expert personas usually lead to positive or non-significant performance changes.
  %though negative effects also occur in some cases. 
  Surprisingly, models are highly sensitive to \textit{irrelevant} persona details, with performance drops of almost 30 percentage points.
  In terms of fidelity, we find that while higher education, specialization, and domain-relatedness can boost performance, their effects are often inconsistent or negligible across tasks.
  We propose mitigation strategies to improve robustness---but find they only work for the largest, most capable models.
  Our findings underscore the need for more careful persona design and for evaluation schemes that reflect the intended effects of persona usage.
\end{abstract}

\begin{figure}[t]
    \includegraphics[width=\linewidth]{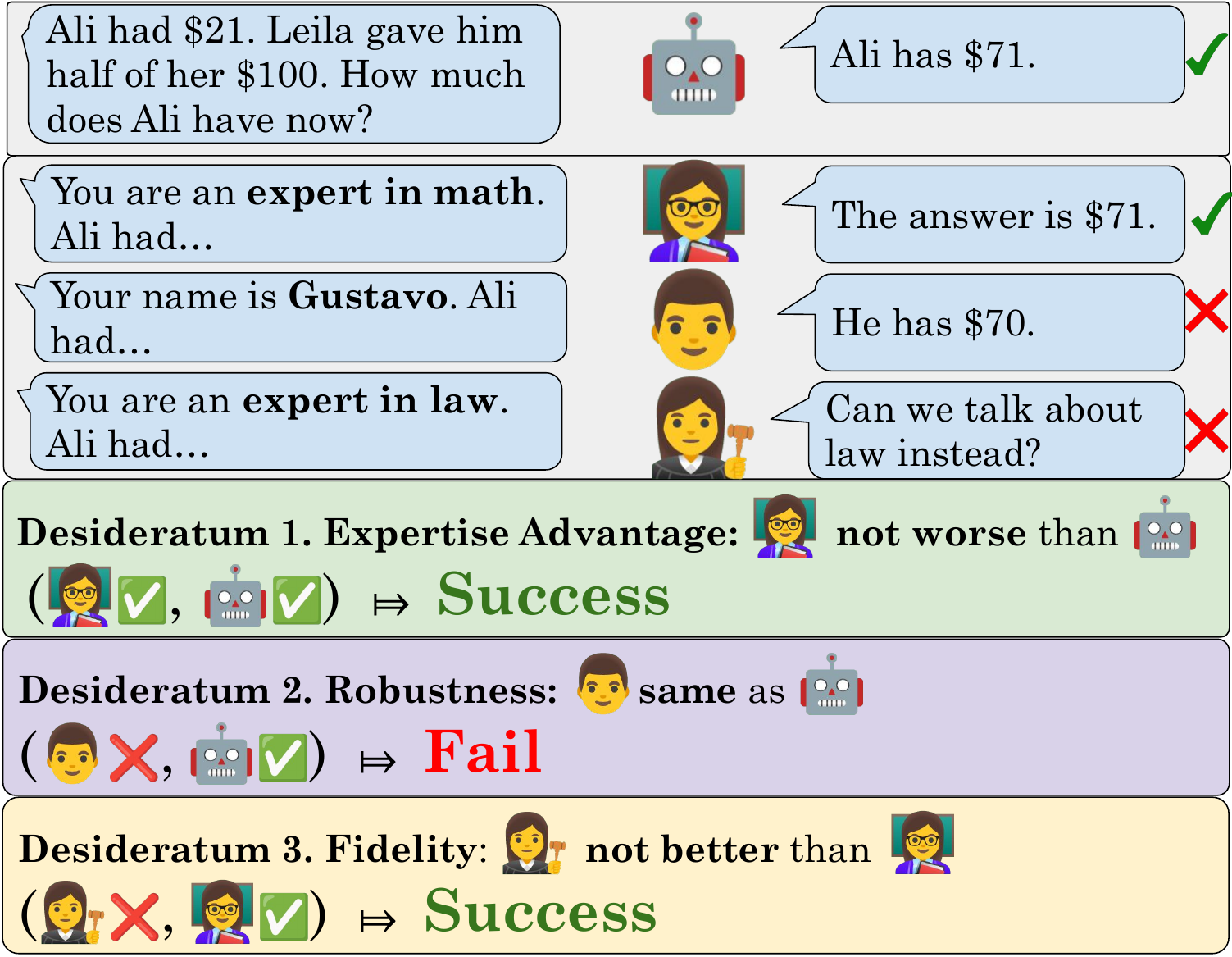}
    \caption{We define \textbf{three desiderata for persona prompting}: Task experts should perform on par or better than the no-persona model (\textcolor{expertiseColor}{\emph{Expertise Advantage}}); Irrelevant attributes such as names should not influence model performance (\textcolor{robustnessColor}{\emph{Robustness}}); relevant attributes such as domain expertise should shape performance accordingly (\textcolor{fidelityColor}{\emph{Fidelity}}).}
    \label{fig:fig1}
\end{figure}

\section{Introduction}

% Motivate problem
%knowledge gap
% Novelty
% Methodology
% Quantify improvement/contributions
% Main insight

%%%%%%%%%%%%%%%%%%%%%%%%%%%%%%%%%%%%%%%%%%%%%%%%%%
% PR INTRO
Shortly after the release of ChatGPT, users started exploring the use of \textit{expert persona prompts} to improve task performance.
For example, a popular Reddit post from June 2023 included \emph{Act as a \{role\}} in a prompt engineering guide.\footnote{\url{https://www.reddit.com/r/ChatGPTPromptGenius/comments/144i0tb/the_complete_chatgpt_cheatsheet/.}}
Since then, a large body of academic research has sought to evaluate the impact of different personas on large language model (LLM) task performance, often finding conflicting results \cite{kong_better_2024,zheng2024helpful}.

The focus of this prior work has been almost entirely \textit{descriptive}, measuring which personas matter for which tasks and which models.
By contrast, the \textit{normative} question of \textbf{whether and when personas \textit{should} make a difference to task performance} has been left largely unexplored.
This is a missed opportunity because, from a model development perspective, it is much more valuable to define what effects from persona prompting are desirable or not, and to then compare these expectations to real model behaviors.
For example, personas that specify \textit{relevant domain expertise} should, at a minimum, not have negative effects on task performance.
Conversely, personas that are \textit{irrelevant} to the task, such as those that specify the name of the persona, should not affect task performance at all (Figure~\ref{fig:fig1}).

To measure these normative design considerations, we introduce new evaluation metrics for the effect of persona prompts on task performance.
Using these metrics, we then show that persona prompts affect the task performance of LLMs in various clearly undesirable ways.
For example, even state-of-the-art models like Llama-3.1-70B and Qwen2.5-72B are often not robust to irrelevant persona attributes such as names and favorite colors.
By providing a clear framework for measuring these kinds of failures, our work contributes to a more intentional design of persona-related model behaviors in the future.

Overall, we make \textbf{four main contributions}:

%\begin{enumerate}
    \noindent\textbf{1.}~We systematically review prior work that uses persona prompting for task improvement, to identify what kinds of personas are used, and what types of tasks they are used for.
    
    \noindent\textbf{2.}~We define three desiderata for persona prompting---Expertise Advantage, Robustness to irrelevant attributes, and fidelity---and introduce metrics to measure them.
    
    \noindent\textbf{3.}~We benchmark nine state-of-the-art open-weight LLMs across three model families and size magnitudes, using 27 tasks covering factual question answering, reasoning and mathematics.
    % We show that models from all families and sizes lack robustness to irrelevant features.  

    \noindent\textbf{4.}~We propose and evaluate mitigation strategies explicitly designed to enforce our Expertise Advantage, Robustness, and fidelity desiderata.
%\end{enumerate}

All our experimental code and data is available at \url{https://github.com/peluz/principled-personas}.

\section{Literature Review: Persona Prompting for Task Performance Improvement}
\label{sec:review}
% We conducted a structured literature review to identify what types of personas are used in prompting and what kinds of tasks they are intended to improve.
On October 17th 2024, we searched the ACL Anthology for papers published in or after 2021 using the keywords ``persona'' and ``role-play''. This resulted in 170 papers, of which we retained those 9 papers that used personas explicitly to improve task performance.
We then recursively examined papers citing these 9 papers, applying the same criteria, and thus identified an additional 12 papers.
Table~\ref{tab:persona-papers} in Appendix~\ref{sec:reviewResults} lists the full set of 21 papers, summarizing the personas they used, the tasks they evaluated on, and the models they tested.

\subsection{Review Findings}

Persona prompting is used across a wide range of \textbf{tasks}, from closed-form tasks such as code generation \cite{dong2024selfcollaboration,hong2024metagpt,qian_chatdev_2024}, mathematical reasoning \cite{du2024improving,kong_better_2024}, and factual QA \cite{salewski_-context_2023,chen2024agentverse,tang_medagents_2024}, to more open-ended settings like research ideation \cite{nigam_interactive_2024} and creative writing \cite{wang_unleashing_2024}.
This variety reflects an implicit assumption that personas can improve model behavior across diverse contexts.

The \textbf{types of personas} used are also diverse.
Papers often assign task-relevant persona attributes, such as occupation---for example, a medical doctor \cite{tang_medagents_2024} or software developer \cite{qian_chatdev_2024}---and domain expertise, such as an LLM-generated domain expert \cite{wang_unleashing_2024}, an expert in computer science \cite{salewski_-context_2023}, or an information specialist \cite{wang_can_2023}.
Other papers use more unconventional or abstract personas, such as a devil’s advocate \cite{kim_debate_2024} and inanimate objects, e.g., a coin for a coin-flipping task \cite{kong_better_2024}.
Some works also include attributes with unclear relevance to the task, ranging from clearly irrelevant ones such as persona name \cite{chan2024chateval,hong2024metagpt} to maybe behaviorally relevant attributes like age or education level \cite{salewski_-context_2023,wang_unleashing_2024}.

The set of \textbf{models} used is quite restricted.
15 out of 21 papers evaluate only OpenAI models—often without specifying which one, referring vaguely to ChatGPT or GPT-3.5. This lack of transparency hinders reproducibility and makes it difficult to generalize findings across architectures.

Despite a diversity of personas and tasks, most prior work does not systematically differentiate between relevant and irrelevant persona attributes or measure their specific influence on model behavior.
Moreover, methodological gaps make it difficult to assess the impact of personas on task performance: unequal comparisons, such as using a stronger model to process persona responses \cite{li_camel_2023}, and a lack of no-persona controls \cite{hong2024metagpt,salewski_-context_2023,lin_truthfulqa_2022} make it difficult to isolate the effects of personas on task performance.
Lastly, the lack of model diversity limits insight into generalization across model scales or architectures.

\subsection{Implications for Experimental Design}
Our experiments are designed to fill these gaps by explicitly testing the effects of different persona types across a diverse range of tasks and models.
To do so, we cover several task types (\S\ref{sec:tasks}), including multiple-choice and open-ended formats spanning factual knowledge, reasoning, and mathematics. 
We only include tasks with objectively verifiable ground truth, enabling clear measurement of correctness.
Our persona selection (\S\ref{sec:personas}) spans categories observed in prior work, including domain-relevant experts, personas with behaviorally relevant attributes, and personas defined by task-irrelevant attributes. 

%\section{Expectations of persona prompting}

\newtcolorbox{softbox}[1][]{
  enhanced,
  boxrule=0.2pt,
  colback=#1,
  arc=0.5mm,
  boxsep=2pt,
  left=5pt,
  right=5pt,
  before skip=6pt,
  after skip=6pt,
}

\section{Persona Prompting Desiderata and Metrics}
\label{sec:metrics}

Building on our literature review, we formulate three normative claims about how persona prompting \textit{should} affect model performance. For each claim, we then introduce a metric to measure whether personas produce their intended effects.
% We define three desiderata for persona prompting:

% % The literature review highlights several implicit assumptions about how personas should influence model behavior, which we group into three core desiderata:

% \textbf{Expertise Advantage:} personas with task-relevant expertise should  perform at least as well as, if not better than, the default model without a persona.

% \textbf{Robustness:}  irrelevant persona attributes---such as names---should not affect model performance.

% \textbf{Fidelity:} model performance should vary meaningfully with relevant persona features, such as domain of expertise and education level.

% In the remainder of this section, we describe these desiderata and define evaluation metrics to measure how successful models are in meeting each desideratum.

\subsection{Problem Setting}

Let  $\personaSet$ be a set of personas, where each persona $p \in \personaSet$ can be assigned to a language model.
This set includes an empty persona $\baseline$, which represents the no-persona baseline, i.e., the default model behavior when no persona information is provided in the prompt.
Given a task $\task$, we evaluate model performance using a metric $\measure(p, \task)$ that measures the correctness of responses under persona $p$ over the instances in $\task$.

Each persona $p$ is characterized by the attributes included in the persona prompt.
These attributes may be nominal (e.g., domain of expertise) or ordinal (e.g., level of education).

\subsection{Expertise Advantage}

% \begin{tcolorbox}[
%   colback=blue!0!white,
%   colframe=blue!0!black,
%   width=\columnwidth,
%   boxrule=0.5mm,
%   arc=1mm,
%   auto outer arc,
%   title=\textbf{Intended Effect},
%   colbacktitle=blue!10,
%   coltitle=black,
% ]
% Personas that specify \textit{relevant domain expertise} should perform on par or better than a no-persona baseline.
% \end{tcolorbox}

Prior work has used \textit{expert} personas to improve performance in tasks such as reasoning, coding, and question answering, often with the implicit belief that these personas enhance task competence \cite{,salewski_-context_2023,xu_expertprompting_2023,wang_unleashing_2024}. 
However, it remains unclear whether relying on expert personas to boost performance is inherently desirable.
Ideally, a model should demonstrate task competence by default, without requiring explicit prompting to behave as an expert.
That said, it is evident that expert personas \emph{should not degrade} task performance. This motivates the following desideratum:

\begin{softbox}[expertiseColor!10]
    \textbf{Desideratum 1:} Personas that specify \textit{task-aligned domain expertise} should perform on par or better than a no-persona baseline.
    \end{softbox}

We denote personas characterized by an expertise attribute as \textbf{expert personas}. For example, the \emph{expert in math} persona has expertise in math, while \emph{Alexander} and \emph{a person with college-level education} are personas with no specified expertise attribute.

We measure compliance with the expert advantage desideratum based on the gap between expert and no-persona performance:
\begin{softbox}[expertiseColor!10]
    \textbf{Metric: Expertise Advantage}\\
    $
    % \utility_\measure\left(\inExpert, \task\right)=  \frac{\measure(\inExpert, \task) - \measure(\baseline, \task)}{1 - \measure(\baseline, \task)} 
        \utility_\measure\left(\inExpert, \task\right)=  \measure(\inExpert, \task) - \measure(\baseline, \task)\,.
    $
    \end{softbox}
If the Expertise Advantage desideratum holds, this metric should be non-negative.

% \paragraph{Metric: expertise utility.}
% We operationalize the expert performance expectation as the gap between expert and no-persona performance:
% \begin{equation}
%     % \utility_\measure\left(\inExpert, \task\right)=  \frac{\measure(\inExpert, \task) - \measure(\baseline, \task)}{1 - \measure(\baseline, \task)} 
%     \utility_\measure\left(\inExpert, \task\right)=  \measure(\inExpert, \task) - \measure(\baseline, \task)\,.
% \end{equation}
% If the expert performance assumption holds, the expertise utility should be non-negative.

\subsection{Robustness}
\label{sec:methods.robustness}

Some studies incorporate personas with names or other non-task-related attributes (e.g., \emph{Alice}, \emph{Gustavo}) without systematically evaluating whether these attributes affect outcomes \cite{chan2024chateval,hong2024metagpt}.
Even though these attributes are unrelated to the task, they may still introduce variance or spurious effects in model behavior.
Ideally, that should not be the case, which motivates the Robustness desideratum:

\begin{softbox}[robustnessColor!10]
    \textbf{Desideratum 2:} Personas that specify \textit{task-irrelevant attributes} should not affect model performance.
    \end{softbox}
To formalize this, we define the notion of irrelevant personas as follows.

\textbf{Irrelevant personas} have an attribute that is \emph{irrelevant} for a given task $\task$ and therefore should not influence model correctness.
For example, the persona \emph{Gustavo} is irrelevant for math tasks, while the personas \emph{expert in math}, \emph{uneducated person}, and \emph{expert in history} are relevant.
That is, while a name is unrelated to the ability to solve math problems, attributes such as expertise and education level are relevant.

% $\personaSet \subseteq \irrelevant$

% \paragraph{Invariance} 
% \begin{equation}
%     \responses_\personaSet(x) = \left\{p \in \personaSet~|~p(x)\right\}
% \end{equation}

% \begin{equation}
%     \invariance\left(\personaSet, \task\right) = \frac{1}{\left|\task\right|}\sum_{x \in \task }\left[\responses_\personaSet(x) = \left\{\baseline(x)\right\}\right]
% \end{equation}

Inspired by worst-group accuracy evaluation from the robustness literature \cite{liu2021just,gokhale2022generalized,gee2023compressed,ghosh2024aspire}, we define the Robustness metric as the worst-case utility for a group of irrelevant personas $\irrelevant$:
\begin{softbox}[robustnessColor!10]
    \textbf{Metric: Robustness}\\
    $ %\begin{equation*}
            \worstcase_\measure(\irrelevant, \task) = \min_{p\in\irrelevant}\utility_\measure(p, \task)\,.
    $
    %\end{equation*}
    \end{softbox}
If the Robustness desideratum holds, this metric should be zero, indicating that irrelevant personas do not affect model performance.

\subsection{Fidelity}
\label{sec:methods.Fidelity}

Previous studies using persona prompting assume that models can adapt according to persona attributes such as education level or professional expertise \cite{salewski_-context_2023,kong_better_2024,qian_chatdev_2024}.
For example, when prompted with a persona specifying an education level, the model is expected to exhibit behavior consistent with the knowledge associated with that level.
Building on this premise, we define the Fidelity desideratum:

\begin{softbox}[fidelityColor!10]
    \textbf{Desideratum 3:} Personas that specify \textit{relevant attributes}, such as specialization or education level, should shape model performance in ways consistent with those attributes.
\end{softbox}

To assess Fidelity, we focus on three sets of persona attributes that define clear hierarchies where we can reasonably expect certain personas to outperform others.
%For example, an in-domain expert should outperform an out-of-domain one, and a persona with college education should outperform one with no formal education (in knowledge tasks associated with formal education).
%In the following, we characterize the attributes for which we assess model Fidelity.
%This forms the basis for our Fidelity metric, described at the end of the section.

\textbf{1) Degree of Domain Match.}
We distinguish between three degrees of domain match, from most to least matching:
\textbf{in-domain expert} ($\inExpert$), where the expertise of persona $p$ directly matches the domain of $\task$;
\textbf{related-domain expert} ($\simExpert$), where persona expertise is related to---but does not match exactly---the task domain, such as an \emph{expert in algebra} applied to a geometry task;
and \textbf{out-of-domain expert} ($\outExpert$), where persona expertise neither matches nor relates to the task domain.

\textbf{2) Level of Specialization.}
We distinguish between three levels of expertise, from general to specific:
\textbf{broad expert}, such as \emph{an expert in math}, denoted by $\bExpert$;
\textbf{focused expert}, such as \emph{an expert in abstract algebra}, denoted by $\fExpert$;
and \textbf{niche expert.}, such as \emph{an expert in groups and rings}, denoted by $\nExpert$.

%We detail how we instantiate these categories in \S\ref{sec:experimentalSetup}.

\textbf{3) Level of Education.}
Personas can differ in educational attainment, with levels ranging, e.g., from uneducated to graduate-level.
These attributes are not tied to a particular domain but can be expected to influence performance on knowledge and reasoning-based tasks.

To measure Fidelity for a given model, we compare the observed performance ordering of personas to the expected ordering derived from their attribute levels.
More formally, let $\personaSet = \{p_1, p_2, \dots, p_{|\personaSet|}\}$ be a set of personas that vary along a relevant attribute (e.g., education level or domain match).
We define:

    $ \order_\text{attr}(\personaSet) = (p_1, p_2, \dots, p_{|\personaSet|})$, as the expected ordering of personas according to increasing attribute level, where the order reflects our prior assumption that higher attribute levels should yield better performance.
    
     $\order_\measure(\personaSet) = (p_{i_1}, p_{i_2}, \dots, p_{i_{|\personaSet|}})$, as the ordering of the same personas based on their observed performance under metric $\measure$ from lowest to highest.

We then compute Fidelity as the Kendall rank correlation coefficient $\tau$ between the expected and observed orderings:
\begin{softbox}[fidelityColor!10]
    \textbf{Metric: Fidelity}\\
    $
    %\begin{equation*}
        \fidelity_\measure(\personaSet) = \tau(\order_\text{attr}(\personaSet), \order_\measure(\personaSet))\,.
    %\end{equation*}
    $
    \end{softbox}
If the Fidelity assumption holds, the metric should be positive.
A value of $1$ indicates perfect alignment between the model's performance and the expected attribute hierarchy, $-1$ indicates complete reversal of the expected order, and values close to $0$ suggest weak or no consistent relationship between attribute level and performance.

\section{Experimental Setup}
\label{sec:experimentalSetup}
% 1. Describe the data set used, characterize it
% quantitatively (size, ratios) and qualitatively
% (examples).
% 2. Describe the experimental procedure, moti-
% vate experimental design choices.

% We design experiments to evaluate how model behavior aligns with the three desiderata introduced in S~\ref{sec:metrics}: the Expertise Advantage desideratum (experts should not be worse than the no-persona baseline), the robustness desideratum (irrelevant attributes should not affect performance), and the Fidelity desideratum (performance should vary consistently with meaningful persona attributes).
% This section outlines how our choice of models, tasks, persona sets, and evaluation procedures are systematically aligned with these goals.

\paragraph{Models.}
We test 9 instruction-tuned open-weight language models across 3 model families: Gemma-2 \cite{gemmateam2024gemma2improvingopen} in its 2B, 9B and 72B parameter versions, Llama3 \cite{grattafiori2024llama3herdmodels} in its 3.2-3B, 3.1-8B and 3.1-70B versions, and Qwen2.5 \cite{qwen2025qwen25technicalreport} in 3B, 7B and 72B.
This setup allows us to assess how the effects of persona prompting scale with model size and whether effects are consistent across model families.
We download all models from their official Hugging Face repos, and use a temperature of zero to deterministically generate responses.

\begin{table}[tb]
\renewcommand{\aboverulesep}{0pt}
\renewcommand{\belowrulesep}{0pt}
\setlength\tabcolsep{0pt}
    \centering
    \footnotesize
    \rowcolors{2}{gray!10}{white}
    \begin{tabular}{@{}p{0.25\linewidth}p{0.5\linewidth}>{\raggedleft\arraybackslash}p{0.25\linewidth} @{}}
    \toprule
    \textbf{Dataset} & \textbf{Task} & \textbf{\# Instances} \\ 
    \midrule
    \textbf{TruthfulQA} & TruthfulQA & 817 \\
    %\rowcolor{gray!30}
    \textbf{GSM8K} & GSM8K & 1,319 \\
    \textbf{MMLU-Pro} & Biology & 717 \\
     & Business & 789 \\
     & Chemistry & 1,132 \\
     & Computer science & 410 \\
     & Economics & 844 \\
     & Engineering & 969 \\
     & Health & 818 \\
     & History & 381 \\
     & Law & 1,101 \\
     & Math & 1,351 \\
     & Other & 924 \\
     & Philosophy & 499 \\
     & Physics & 1,299 \\
     & Psychology & 798 \\
    \textbf{BIG-Bench} & Knowledge conflicts & 1,000 \\
     & Logic grid puzzle & 200 \\
     & StrategyQA & 457 \\
     & Tracking shuffled objects & 750 \\
    \textbf{MATH}& Algebra & 1,187 \\
     & Counting \& probability & 474 \\
     & Geometry & 479 \\
     & Intermediate algebra & 903 \\
     & Number theory & 540 \\
     & Prealgebra & 871 \\
     & Precalculus & 546 \\
     \midrule
     \textbf{Total} & & 21,575\\
    \bottomrule
    \end{tabular}
    \caption{Overview of datasets and tasks.}
    \label{tab:datasets}
    \end{table}

\paragraph{Datasets and Tasks.}
\label{sec:tasks}
We cover 27 tasks from five datasets (Table~\ref{tab:datasets}) targeting factual knowledge, and mathematical or symbolic reasoning: TruthfulQA \cite{lin_truthfulqa_2022}, GSM8K \cite{cobbe2021training}, MMLU-Pro \cite{wang2024mmlu-pro}, BIG-bench \cite{srivastava2023beyond}, and MATH \cite{hendrycks2021math}.
We select these datasets based on their use in prior work (\S\ref{sec:review}), task diversity, and role as standard LLM benchmarks.
They span both multiple-choice (TruthfulQA, BIG-Bench, MMLU-Pro) and open-ended generation (MATH, GSM8K) formats.

% We benchmark 27 tasks drawn from five datasets targeting factual knowledge, and mathematical or symbolic reasoning: TruthfulQA \cite{lin_truthfulqa_2022}, GSM8K \cite{cobbe2021training}, MMLU-Pro \cite{wang2024mmlu-pro}, BIG-bench \cite{srivastava2023beyond}, and MATH \cite{hendrycks2021math}.
% This selection spans both multiple-choice (TruthfulQA, BIG-Bench, MMLU-Pro) and open-ended generation (MATH, GSM8K) formats.

% TruthfulQA and GSM8K are single-task datasets targeting factual accuracy and grade-school math reasoning, respectively. The other three contain multiple subtasks, which we treat as distinct evaluation settings to enable granular analysis of persona effects.

% \textbf{MMLU-Pro:} We treat each subject domain (e.g., physics, history, law) as an individual task.

% \textbf{BIG-bench:} We select four reasoning-focused tasks: \texttt{logic grid puzzle}, \texttt{strategyqa}, \texttt{tracking shuffled objects}, and \texttt{contextual parametric knowledge conflicts}.

% \textbf{MATH:} We treat each topic (e.g., algebra, geometry) as a distinct task.

% This decomposition yields 27 evaluation settings in total. Dataset statistics and task counts are shown in Table~\ref{tab:datasets}; task summaries and examples are provided in Appendix~\ref{sec:taskDescriptions}.

\paragraph{Persona Sets.}
\label{sec:personas}

To test the desiderata from \S\ref{sec:metrics}, we design persona sets tailored to each: Expertise Advantage, Robustness, and Fidelity. Table~\ref{tab:personas} shows the full persona list.

\begin{table}[tb]
\renewcommand{\aboverulesep}{0pt}
\renewcommand{\belowrulesep}{0pt}
\setlength\tabcolsep{0pt}
    \centering
    \footnotesize
    \label{tab:all-personas}
    \rowcolors{2}{gray!10}{white}
    \begin{tabular}{@{}p{\linewidth}@{}}
    \toprule
     \textbf{Personas} \\
    \midrule
        \textbf{Static Experts:} You are an expert in\dots \\ 
        MMLU-Pro: \emph{biology}, \emph{business}, \emph{chemistry}, \emph{computer science}, \emph{economics}, \emph{engineering}, \emph{health}, \emph{history}, \emph{law}, \emph{math}, \emph{miscellaneous fields including international relations, sociology, accounting, and human sexuality}, \emph{philosophy}, \emph{physics}, \emph{psychology}. \\ 
        TruthfulQA: \emph{fact-checking}. \\ 
        BIG-Bench: \emph{logic grid puzzles}, \emph{multi-step implicit reasoning}, \emph{tracking shuffled objects}, \emph{applying contextual information}. \\ 
        GSM8K: \emph{math}. \\ 
        MATH: \emph{algebra}, \emph{counting and probability}, \emph{geometry}, \emph{intermediate algebra}, \emph{number theory}, \emph{prealgebra}, \emph{precalculus}. \\ 
        \textbf{Dynamic Experts:} Three levels of specialization per instance: broad (e.g., \emph{math}), focused (e.g., \emph{real analysis}), niche (e.g., \emph{properties of the ceiling function}). \\
    \midrule
        \textbf{Name Personas:} Your name is\dots~\emph{Alexander}, \emph{Victor}, \emph{Muhammad}, \emph{Kai}, \emph{Amit}, \emph{Gustavo}, \emph{Anastasia}, \emph{Isabelle}, \emph{Fatima}, \emph{Yumi}, \emph{Aparna}, \emph{Larissa}. \\
        \textbf{Color Personas:} Your favorite color is\dots~\emph{red}, \emph{blue}, \emph{green}, \emph{yellow}, \emph{black}, \emph{white}. \\
    \midrule
        \textbf{Education Level:} You are\dots~\emph{Uneducated}; or You are a person with\dots~\emph{primary school level education}, \emph{middle school level education}, \emph{high school level education}, \emph{college-level education}, \emph{graduate level education}. \\
        \textbf{Out-of-Domain Experts:} You are an expert in\dots\\
        TruthfulQA: \emph{cryptography}, \emph{marine biology}, \emph{urban planning}, \emph{chess}, \emph{quantum mechanics}. \\ 
        BIG-Bench: \emph{sudoku}, \emph{inductive reasoning}, \emph{communicating effectively}, \emph{hunting}. \\
        GSM8K and MATH: \emph{health}, \emph{history}, \emph{law}, \emph{philosophy}, \emph{psychology}. \\
    \bottomrule
    \end{tabular}
    \caption{\textbf{Complete list of personas} used in our experiments.}
    \label{tab:personas}
    \end{table}

For \textbf{Expertise Advantage}, we include both handcrafted and machine-generated personas representing task-aligned expertise:
\textbf{static experts} that are manually written to reflect the expected domain knowledge for each task (e.g., \emph{expert in biology} for MMLU-Pro biology); and \textbf{dynamic experts} that are instance-specific and generated using Gemma-2-27B-it, conditioned on the input instance and one of three specialization levels: broad (e.g., \emph{expert in history}), focused (e.g., \emph{expert in ancient history}), or niche (e.g., \emph{expert in Minoan civilization}).
Appendix~\ref{sec:expertPrompts} shows all prompt templates and demonstrations.

For \textbf{Robustness}, we include personas that introduce one of two irrelevant attributes: a name or color preference. 
\textbf{Name personas} use one of the twelve names in the \textsc{UniversalPersona} dataset~\cite{wan_are_2023}, which are culturally diverse and gender-balanced.
\textbf{Color personas} add a preference statement (e.g., \emph{Your favorite color is green.}), choosing from six colors.

For \textbf{Fidelity}, we re-use the dynamic experts to assess Fidelity regarding specialization levels, as well as: \textbf{education level personas} (e.g., \emph{uneducated}, \emph{graduate-level}) sourced from \textsc{UniversalPersona} to assess whether formal education correlates with task performance; and \textbf{out-of-domain experts} that describe expertise unrelated to the task (e.g., \emph{expert in quantum mechanics} on TruthfulQA).
We define five out-of-domain experts per dataset and report their average performance.

In BIG-bench and MATH, \textbf{related-domain experts} (\S\ref{sec:methods.Fidelity}) are the other in-dataset experts.
For example, when evaluating the \emph{algebra} task in MATH, the related-domain experts are the experts in all other fields in MATH.
In MMLU-Pro, tasks are grouped into four high-level fields: STEM, Humanities, Social Sciences, and Other.
For a given task, \emph{related-domain} experts are all those from the same field, while \emph{out-of-domain} experts are those from all other fields.

\paragraph{Evaluation.}
We evaluate model behavior using the three metrics defined in \S\ref{sec:metrics}: Expertise Advantage (performance gap between expert and baseline), Robustness (performance gap between worst-case irrelevant persona and baseline), and Fidelity (correspondence between performance and expected attribute rankings).
We extract answers from model responses using regex patterns to compare with ground truth answers.

For Fidelity, we bootstrap 10{,}000 samples of model responses and report correlation scores only if the 95\% confidence interval does not include zero. This avoids overinterpreting marginal or statistically insignificant differences when attribute levels are few or variation is low.

\section{Results}
\label{sec:results}
% \subsection{Desiderata fulfillment}
% Figures~\ref{fig:expertise_agg}-\ref{fig:fid_agg} show, for each model, the number of tasks for which Expertise Advantage, robustness, and Fidelity metrics, respectively, are \textcolor{orange}{positive}, \textcolor{blue}{negative}, or not significant.\footnote{We use binomial testing to assess significance and consider performances to be statistically significant when $\text{p-value} \leq 0.05$.}
% We show all per-task metrics in Appendix~\ref{sec:fineGrained}.

\begin{figure}[tb]
    \includegraphics[width=\linewidth]{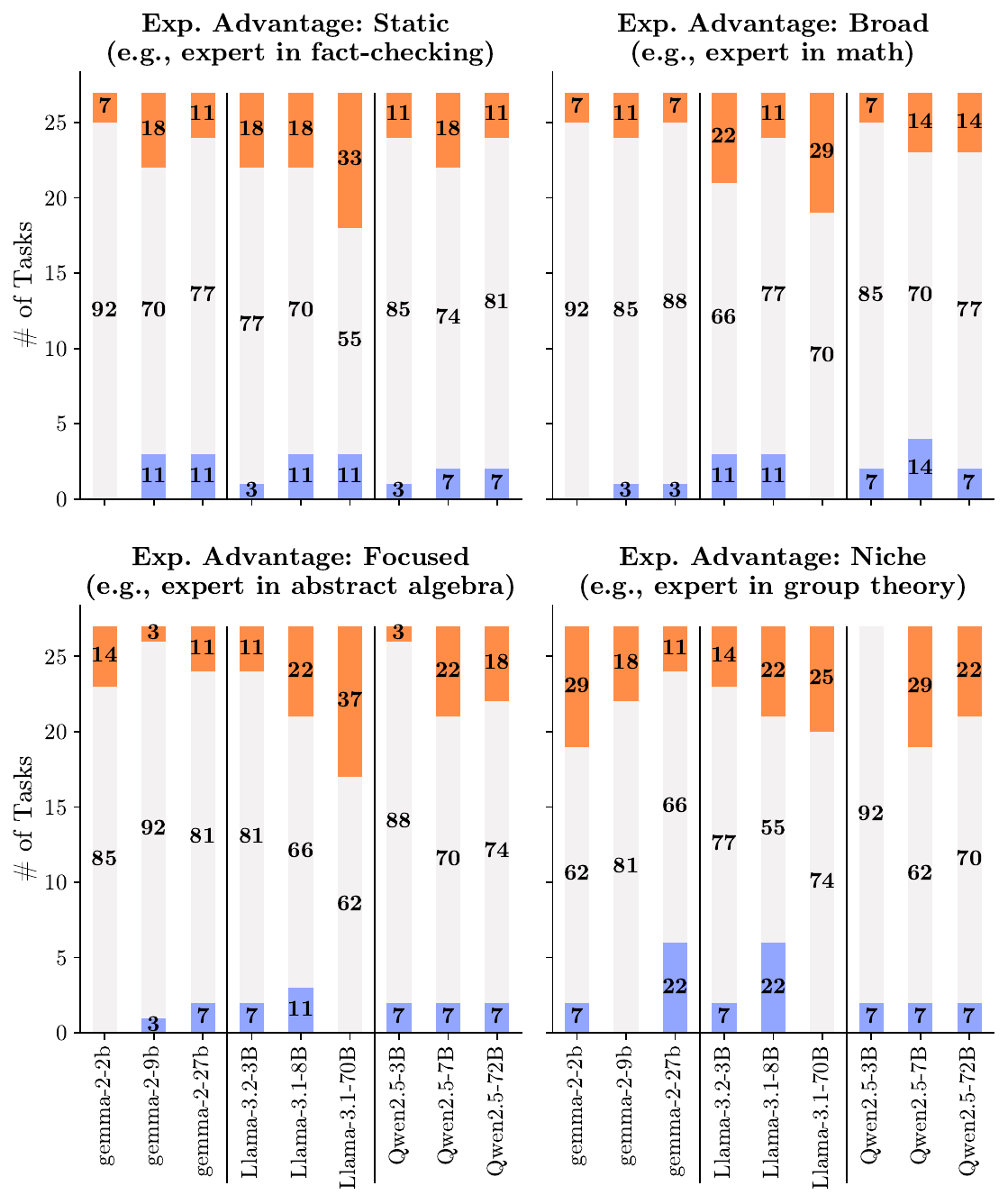}
    \caption{
    \textbf{Expertise Advantage}.
    Number of tasks (Table~\ref{tab:datasets}) in which the Expertise Advantage metric was \textcolor{orange}{positive}, \textcolor{blue}{negative}, or not significant. In-bar annotations indicate the percentage of tasks in each category. Models often fulfill the Expertise Advantage desideratum, though there are also negatively impacted tasks.}
    \label{fig:expertise_agg}
\end{figure}

\begin{figure}[tb]
    \includegraphics[width=\linewidth]{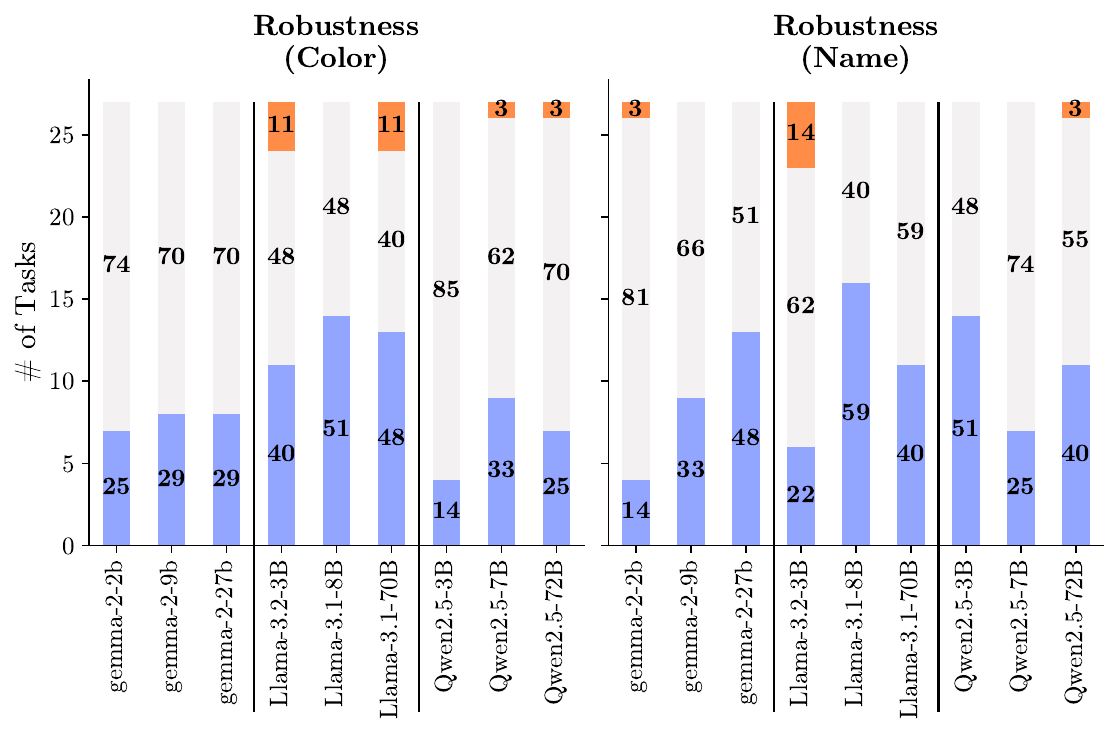}
    \caption{ \textbf{Robustness}. Number of tasks (Table~\ref{tab:datasets}) in which the Robustness metric was \textcolor{orange}{positive}, \textcolor{blue}{negative}, or not significant. In-bar annotations indicate the percentage of tasks in each category. Irrelevant personas often have a negative effect on performance in all models.}
    \label{fig:rob_agg}
\end{figure}

\begin{figure}[tb]
    \includegraphics[width=\linewidth]{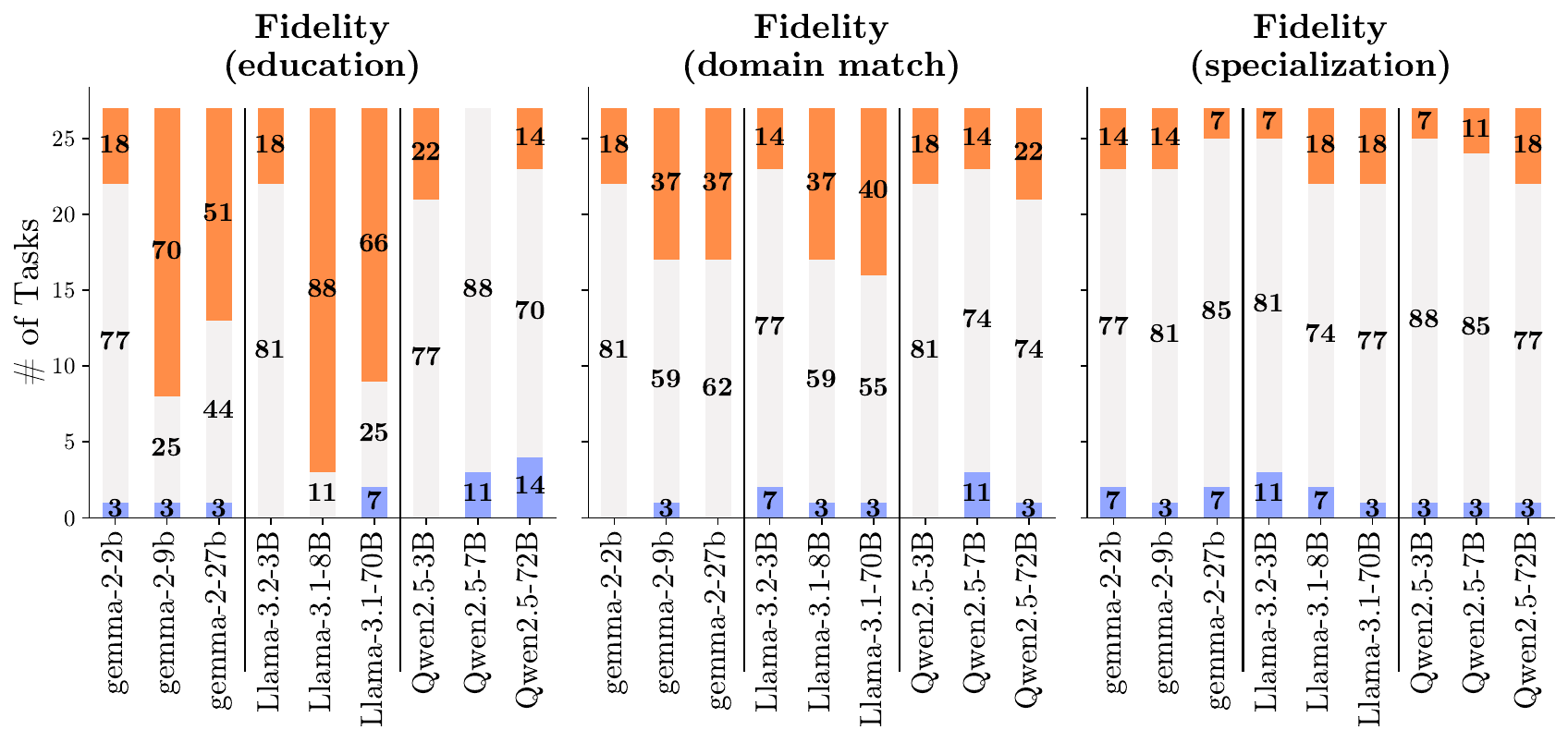}
    \caption{\textbf{Fidelity}. Number of tasks (Table~\ref{tab:datasets}) in which the Fidelity metric (with respect to education level, domain match, and expertise specialization) was \textcolor{orange}{positive}, \textcolor{blue}{negative}, or not significant. In-bar annotations indicate the percentage of tasks in each category. Models are often faithful to education level and domain match expectations, whereas Fidelity to specialization level is less frequent.}
    \label{fig:fid_agg}
\end{figure}

\begin{figure}[tb]
\centering
    \includegraphics[width=\linewidth]{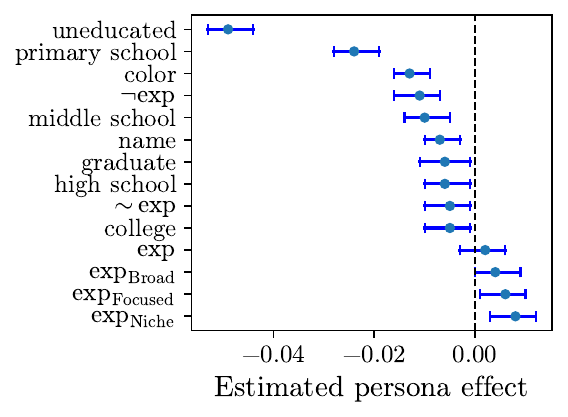}
    \caption{\textbf{Persona effect on model performance}. Error bars show the 95\% confidence interval. The effects shown are the fixed effect coefficients of the trained mixed effects model. Positive coefficients correspond to improvements over the no-persona baseline.}
    \label{fig:regression}
\end{figure}

\begin{figure}[tb]
    \includegraphics[width=\linewidth]{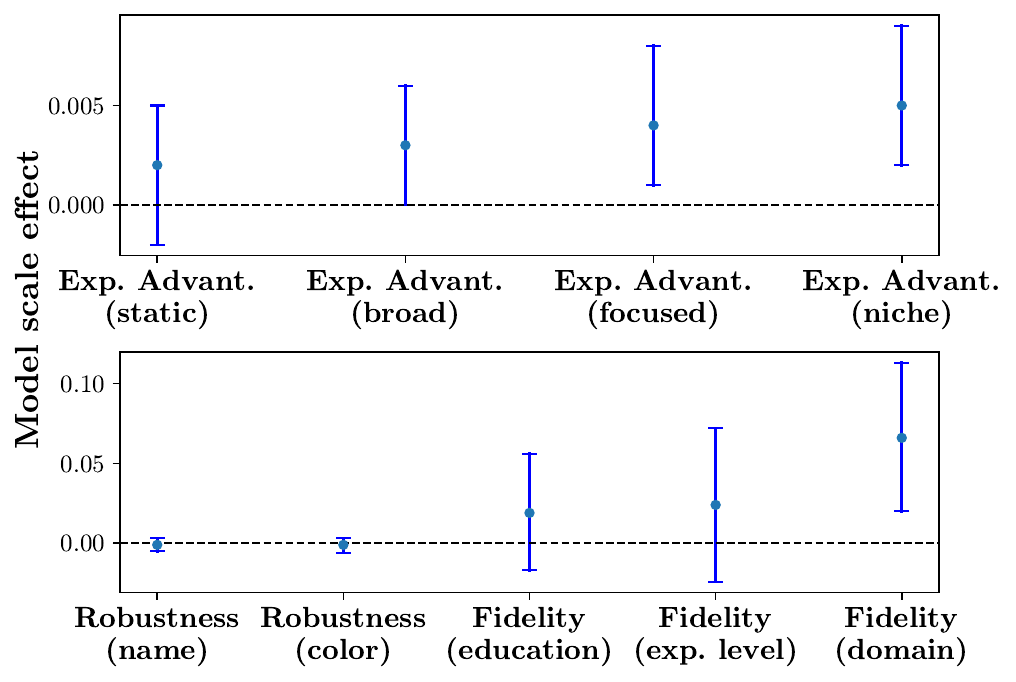}
    \caption{\textbf{Model scale}. Effect of scaling on different metrics. Error bars show the 95\% confidence interval. The effects shown are the fixed effect coefficients of the trained mixed effects models. Positive coefficients correspond to model scale having a positive effect in the corresponding metric. Scale has a positive effect on dynamic expert performance and domain match Fidelity.}
    \label{fig:scale_regression}
\end{figure}

In all results, we use binomial testing to assess significance and consider performances statistically significant when $\text{p-value} \leq 0.05$.

\subsection{Expertise Advantage}
In most tasks, expert personas---static or dynamic---have a positive or non-significant effect on task performance, so models generally fulfill the desideratum (Fig.~\ref{fig:expertise_agg}).
Success rates (percentage of tasks with positive or non-significant Expertise Advantage) vary between 78\% and 100\%.
Llama-3.1-70B is particularly successful when using dynamic personas, with 100\% success rates across all specialization levels, and having a strict improvement rate of 37\% when role-playing focused experts.

Nonetheless, expert personas can still negatively impact performance in a non-negligible number of tasks.
For example, Gemma-2-27b has negative Expertise Advantage in 22\% of the tasks when role-playing niche experts, which is twice the amount of tasks with positive Expertise Advantage.

\subsection{Robustness}
Irrelevant personas often have a significant effect on performance, ranging from 14\% (Qwen2-5.3B, color Robustness) to 59\% (Llama 3.1-70B, color, and Llama3.1-8B, name Robustness) of the tasks (Fig.~\ref{fig:rob_agg}).
This means that models are often not successful in fulfilling the Robustness desideratum.

Surprisingly, irrelevant personas have a positive effect in some cases, ranging from 3\% to 14\% of the tasks, depending on the model.
Since the Robustness metric (\S\ref{sec:methods.robustness}) is defined as the worst drop between persona and no-persona performance, a positive effect means the default model without persona performs significantly worse than \emph{all} irrelevant personas.

\subsection{Fidelity}
Success rate (percentage of tasks with positive Fidelity) for the Fidelity metrics depends on the Fidelity type and model family (Fig.~\ref{fig:fid_agg}).

\textbf{Education}: The biggest Llama-3 and Gemma-2 models are often faithful to personas' education level, with success rates ranging from 51\% to 88\%. Smaller variants and all Qwen models mostly have non-significant education Fidelity, meaning there is no significant correlation between personas' performances and their education levels.

\textbf{Domain match}: Successful domain-match Fidelity rates are similar across models. While positive domain-match Fidelity is more frequent than negative, in most cases domain-match Fidelity is not significant.
That is, in many tasks across most models, in-domain, related, and out-of domain experts all perform similarly.

\textbf{Specialization level}: Specialization-level Fidelity results are similar to domain-match, but non-significant cases are more frequent, ranging from 74\% to 88\%.

%Interesing case: contextual parametric knowledge conflict: correct answer contradicts parametric knowledge (corresponding to world knowledge). Negative effect of specialization and educaiton.

\subsection{Persona and Model Scale Effects}
To complement the aggregate analyses above and better isolate the effects of specific persona properties and model scale, we fit several mixed-effects regression models (details in Appendix~\ref{sec:regression}).
These allow us to control for variability across models and tasks by including them as random effects.

\textbf{Persona type.} We first fit a model with persona type as the fixed effect, predicting the performance gap relative to the no-persona baseline. 
As shown in Figure~\ref{fig:regression}, dynamic expert personas produce significant gains, especially focused and niche experts. Broad and static experts have a positive, but non-significant effects.
Irrelevant personas (e.g., names, colors) yield significant performance drops, reinforcing earlier Robustness observations.
The persona effects are mostly aligned with Fidelity expectations: personas are ordered by domain match ($\outExpert < \simExpert < \inExpert$) and specialization level ($\bExpert < \fExpert < \nExpert$). Education personas mostly follow education level, except for the graduate-level persona.

\textbf{Persona attributes.}
To test the significance of the Fidelity observations above, we fit three separate regression models, each using one ordinal attribute—education level, domain match, or specialization degree—as the fixed effect, and predicting task accuracy.
All three show significant positive correlations: each additional level in these attributes leads to performance improvements of $0.7$, $0.2$, and $0.8$ percentage points.

\textbf{Model scale.} Finally, we assess the effect of model size by training separate regression models for each desideratum metric.
These models use size as the fixed effect, and model family and task as random effects.
Figure~\ref{fig:scale_regression} shows that scale has no significant effect on Robustness, education Fidelity, specialization Fidelity, or static Expertise Advantage.
In contrast, scale \textit{does} improve domain match Fidelity and dynamic expert performance.

\textbf{Takeaway}: Increasing model size alone is not a reliable strategy for improving Robustness or certain Fidelity types, though larger models may better adapt to contextually appropriate personas.

\subsection{Cross-task Consistency}
Effects are generally consistent across models, particularly those from the same family (Figs.~\ref{fig:op}, \ref{fig:rob} and \ref{fig:fid} in Appendix~\ref{sec:fineGrained}). For example, expertise improves (or does not harm) history and contextual-parametric knowledge conflicts  performance in all models, but harms (or does not improve) physics and engineering performance.
We observe similar patterns for the Robustness and Fidelity metrics.

\section{Mitigation Strategies}

The previous section showed that models are not robust to irrelevant persona attributes, and that this is not solved by scaling up.
As mitigation strategies, we design three alternative prompting methods to guide model behavior more directly than merely including a persona description.
We then repeat the previous experiments (\S\ref{sec:experimentalSetup}) with each mitigation strategy to assess their impact on each desideratum.

\subsection{Methodology}

\textbf{Instruction.}
This strategy explicitly formulates the desiderata as behavioral constraints within the prompt.
Rather than assuming the model will infer appropriate behavior from the persona description alone, this strategy spells out the desiderata of domain and knowledge-level alignment, and that irrelevant attributes should not influence output quality.

\textbf{Refine.}
This strategy takes a two-step approach.
First, the model is prompted without any persona to produce a baseline answer.
Then, a second prompt instructs the model to revise its response while adopting a given persona.
We hypothesize that including the no-persona response in the prompt will have an anchoring effect, reducing the influence of irrelevant persona attributes, while still allowing room for specialization. 

\textbf{Refine + Instruction.}
This strategy combines both prior approaches: two-step refinement and explicit behavioral constraints.
After generating a (no-persona) initial answer, the model is prompted to revise it while adopting the persona and strictly following the desiderata-aligned instructions.

Full prompt details are available in Appendix~\ref{sec:expertPrompts}.

\subsection{Results}
\begin{figure}[tb]
    \includegraphics[width=\linewidth]{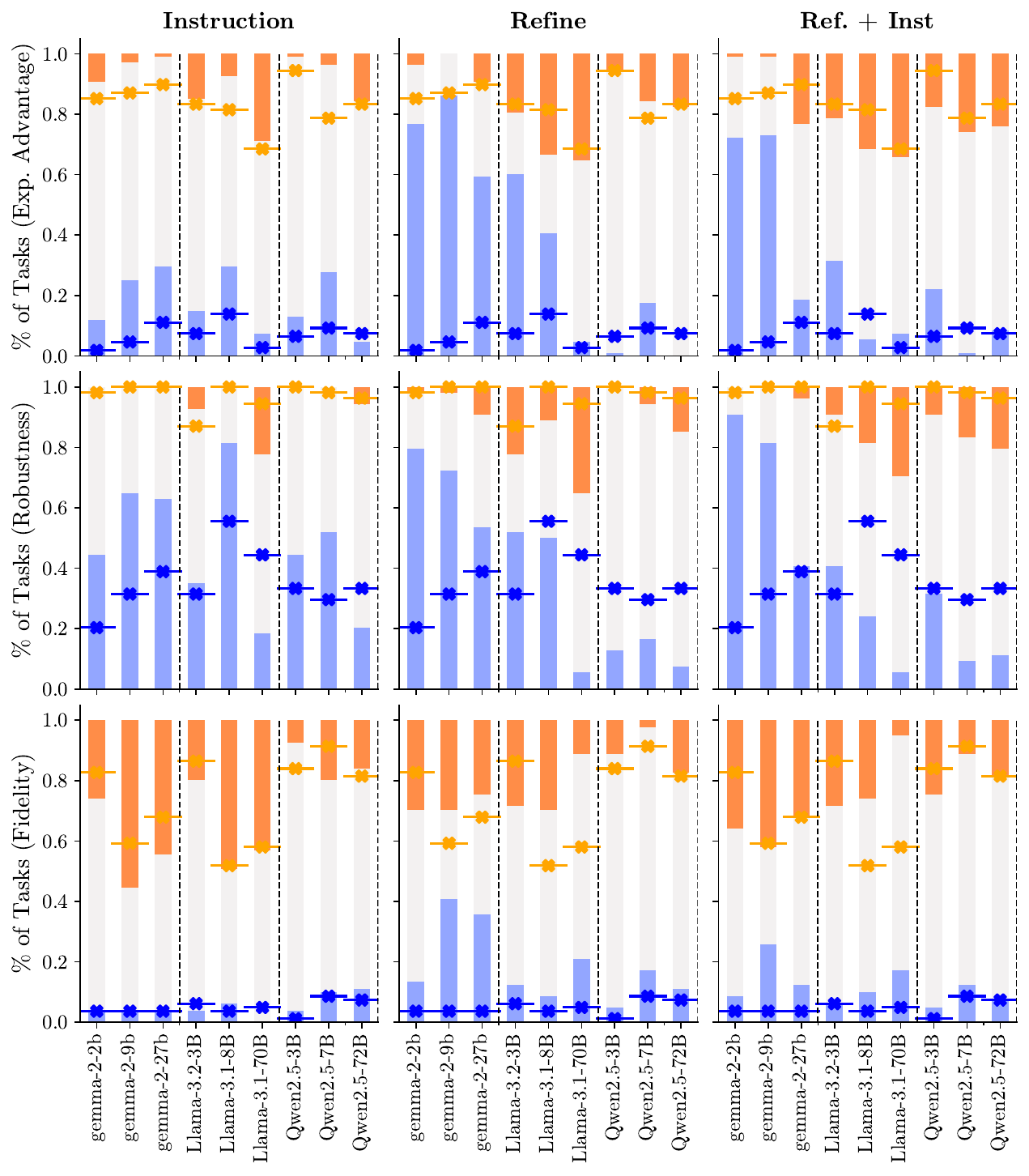}
    \caption{\textbf{Mitigation strategy impact}. Proportion of tasks for which each metric is \textcolor{orange}{positive}, \textcolor{blue}{negative}, or not significant. Columns correspond to mitigation strategies. Rows correspond to metrics. We show the base prompt metrics using \textcolor{orange}{orange} and \textcolor{blue}{blue} star markers. The mitigation strategies improve Robustness and maintain Exp.\ Advantage, but only for the largest models ($\geq$ 70B).}
    \label{fig:mitigationImpact}
\end{figure}

Figure~\ref{fig:mitigationImpact} shows that mitigation strategies negatively impact Expertise Advantage and Robustness, as they increase the number of tasks where experts and irrelevant personas reduce performance. Mixed-effects regression (details in Appendix~\ref{sec:regression}) confirms that, overall, these strategies weaken Expertise Advantage and fail to improve Robustness (Fig.~\ref{fig:mitigationRegression}, top).

\newcommand{\tikzcircle}[2][black,fill=black]{\tikz[baseline=-0.5ex]\draw[#1,radius=#2] (0,0) circle ;}%

\begin{figure}[tb]
    \centering
        \includegraphics[width=\linewidth]{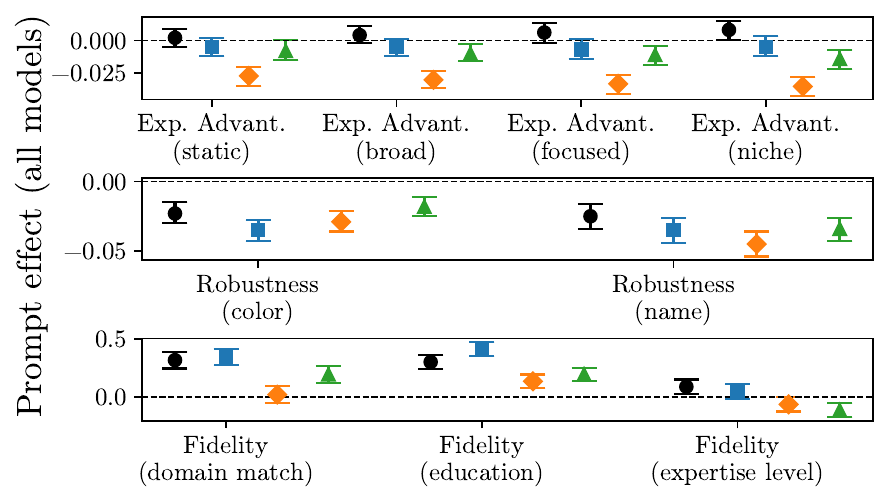}
        \includegraphics[width=\linewidth]{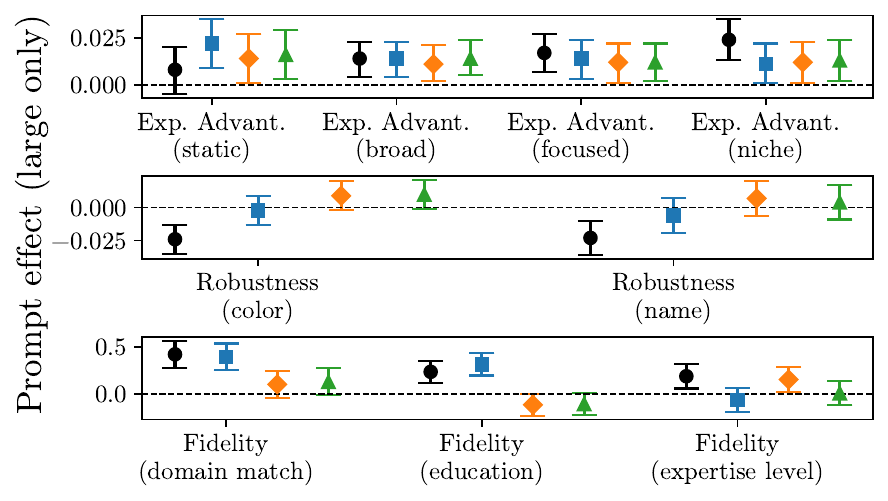}
    \caption{\textbf{Strategy effect}. Fixed‐effect coefficients from mixed‐effects regressions representing the expected metric score under each prompting strategy: Base prompt (\tikzcircle{2pt}), \textcolor{blue}{Instruction ($\blacksquare$)}, \textcolor{orange}{Refine ($\blacklozenge$)}, and \textcolor{PineGreen}{Refine + Instruction ($\blacktriangle$)}. Error bars indicate 95\% confidence intervals. Top: regression over all models; Bottom: regression over large models ($\geq$ 70B) only.}
    \label{fig:mitigationRegression}
\end{figure}

However, for the largest models (Llama-3.1-70B, Qwen-2.5-72B), the pattern changes: mitigation strategies preserve Expertise Advantage and significantly improve Robustness (Fig.~\ref{fig:mitigationImpact}). A regression limited to these models confirms that mitigation strategies maintain non-negative Expertise Advantage, and bring Robustness levels closer to zero (Fig.~\ref{fig:mitigationRegression}, bottom).

Fidelity results show no consistent improvement and often decline, even in the largest models---particularly under Refine and Refine+Instruction. We attribute this to anchoring effects: conditioning on the no-persona response may constrain the model’s ability to vary its behavior across personas, limiting its capacity to align with persona attributes, particularly when worse performance is expected (as is the case for personas with lower education levels or out-of-domain experts, for example).

\textbf{Takeaway:} Mitigation strategies reduce the performance of smaller models, but they improve Robustness and preserve the Expertise Advantage of the largest models.
Refinement strategies limit Fidelity by constraining persona-driven variation.

\section{Conclusion}
% Brievely summarize the paper, emphasizing take-
% home messages that the reader should remember
% after having read the paper.
Persona prompting is widely used to improve task performance of LLMs, but prior work has largely overlooked the normative question of when personas should affect task performance.
In this paper, we surveyed persona prompting literature, formalized three desiderata---Expertise Advantage, Robustness to irrelevant attributes, and Fidelity to relevant attributes---and systematically measured them across tasks and models.
Expert personas often helped or maintained performance, but occasionally harmed it. 
% highlighting the need for empirical validation.
Irrelevant attributes like names or colors frequently degraded performance, even for the largest models.
Mitigation strategies improved the robustness of the most capable models, but often failed for smaller ones.
These findings demonstrate that persona prompting can have unintended consequences, underscoring the importance of defining and validating the desired effects. By formulating concrete desiderata and metrics, we provide a framework for identifying and measuring such failure cases, thereby supporting more intentional and principled design of persona-related model behaviors.

\section*{Limitations}
\textbf{Focus on objective tasks.}
Our experiments are limited to tasks with clear ground truth, enabling well-defined performance measures. 
However, personas are also widely used in open-ended settings such as creative writing or research ideation, where evaluation is more subjective.
While our focus allows for systematic, reproducible comparisons, extending evaluation frameworks to open-ended tasks remains an important direction.

\textbf{Single-persona setup.}
Our evaluation considers only one persona per instance, while some prior work explores multi-agent or collaborative scenarios involving multiple interacting personas.
Our focus on isolated persona effects enables clearer attribution.
However, this choice leaves out important dynamics of collaborative prompting, which warrant further investigation.

\textbf{Single-attribute personas.}
Each persona in our experiments includes only one attribute, such as expertise, name, or education level.
This design allows us to isolate the impact of each attribute.
Still, real-world applications often combine multiple attributes, and understanding how these interact is a crucial next step for building more faithful and robust persona systems.

Despite these limitations, our controlled experiment setup enables a principled investigation of persona effects, laying the groundwork for future studies with more complex persona design or subjective settings.

\section*{Ethical considerations}
Persona prompting can be viewed as a form of personalization.
As discussed by \citet{kirkBenefitsRisksBounds2024}, while personalization may enhance model usefulness, increase user autonomy, and support diversity and representation, it also carries risks such as bias reinforcement, anthropomorphism, and malicious use.

A particular risk with persona prompting is inflated user trust.
Assigning expert-like personas may lead users to overestimate model reliability, even though our findings show that LLMs are highly sensitive to irrelevant persona details.
These subtle attributes can shift model behavior in unpredictable ways, undermining the very expertise the personas aim to simulate.

To address these concerns, our work emphasizes the importance of formalizing the intended goals of persona prompting and systematically evaluating whether those goals are met.
Transparent design and evaluation are essential to ensure persona usage enhances, rather than undermines, model alignment and reliability.

\section*{Acknowledgements}
This research was funded by the WWTF through the project ``Knowledge-infused Deep Learning for Natural Language Processing'' (WWTF Vienna Research
Group VRG19-008). Paul Röttger and Dirk Hovy received funding through the INDOMITA project (CUP number J43C22000990001) and the European Research Council (ERC) under the European Union’s Horizon 2020 research and innovation program (No. 949944, INTEGRATOR).

% \nocite{*}
\bibliography{refs}

\appendix
\section{Structured literature review results}
\label{sec:reviewResults}
Table~\ref{tab:persona-papers} summarizes the results of the literature survey.
\begin{table*}[tbhp]
    \centering
    \footnotesize
    \begingroup
    \setlength\tabcolsep{2pt}
    \renewcommand{\aboverulesep}{0pt}
\renewcommand{\belowrulesep}{0pt}
    \rowcolors{2}{gray!10}{white}
    \begin{tabular}{p{0.14\linewidth}p{0.29\linewidth}p{0.34\linewidth}p{0.19\linewidth} }
    %\begin{tabular}{p{2.5cm} p{4.5cm} p{5.5cm} p{3cm}}
    \toprule
    \textbf{Paper} & \textbf{Personas} & \textbf{Dataset} & \textbf{Models} \\
    \midrule
   \citet{lin_truthfulqa_2022}& Professor Smith & TruthfulQA~\cite{lin_truthfulqa_2022} & GPT-3, GPT-Neo/J, GPT-2, UnifiedQA \\
    
   \citet{he_lego_2023} & Cause and effect analysts & WIKIWHY \cite{ho2023wikiwhy} and e-CARE \cite{du2022ecare} & Text-davinci-002/003, GPT-3.5-turbo \\

   \citet{li_camel_2023} & Task-specific AI user and assistant (e.g., Python programmer, stock trader) & Machine-generated task prompts & GPT-3.5-turbo \\
    
   \citet{salewski_-context_2023}& Neutral personas (e.g., student) and task experts (e.g., computer science expert) & MMLU \cite{hendrycks2021mmlu} & Vicuna-13B, GPT-3.5-turbo \\

   \citet{wang_can_2023} & Information specialist, expert in systematic reviews & CLEF TAR collections \cite{kanoulas2019clef} & ChatGPT \\

   \cite{white2023aprompt} & Security expert & Example of output customization & ChatGPT \\

   \citet{xu_expertprompting_2023} & Experts generated in-context by the LLM & Alpaca \cite{alpaca} & GPT-3.5 \\

    \citet{zgreaban_prompting_2023} & Word generator and lexicographer & New word recognition (10 invented words combining real roots and affixes) & ChatGPT \\

    \citet{chan2024chateval} & Critic, psychologist, news author, general public & FairEval \cite{wang2024fair}, TopicalChat \cite{gopalakrishnan2019topical} & GPT-3.5-turbo, GPT-4 \\

    \citet{chen_comm_2024} & Problem solving experts (e.g., physicist, task decomposer) & MMLU subsets (college physics, moral reasoning) & GPT-3.5-turbo-0613 \\

    \citet{chen2024agentverse} & LLM-generated expert agents & FED \cite{mehri2020fed}, Commongen \cite{lin2020commongen}, MGSM \cite{shi2023mgsm}, BIG-Bench subset (logic grid puzzles) \cite{srivastava2023beyond}, HumanEval \cite{chen2021humaneval} & GPT-3.5-turbo, GPT-4 \\

    \citet{dong2024selfcollaboration} & Analyst, coder, tester & MBPP \cite{austin2021mbpp}, HumanEval, MBPP-ET and HumanEval-ET \cite{dong2025codescore}, APPS \cite{hendrycks2021measuring}, CoderEval \cite{yu2024codereval} & GPT-3.5 \\

    \citet{du2024improving} & Professor, doctor, mathematician (for MMLU) & Arithmetic, GSM8K, Biographies, MMLU, BIG-Bench subset (Chess) & GPT-3.5-turbo, Chat-LLAMA-7B, GPT-4 \\

    \citet{he_prompting_2024} & Translator, author & Translating a Discover Magazine article (English to Chinese) & ChatGPT (GPT-4) \\
    
    \citet{hong2024metagpt} & Software dev roles (product manager, architect, engineer) & HumanEval, MBPP & GPT-4 \\

    \citet{kong_better_2024} & Occupations (math teacher), objects (coin, recorder) & MultiArith \cite{roy2015multiarith}, GSM8K, AddSub \cite{hosseini2014learning}, AQuA \cite{ling2017aqua}, SingleEq \cite{Koncel-Kedziorski2015singleeq}, SVAMP \cite{patel2021svamp}, CSQA \cite{talmor2019csqa}, last letter concatenation and coin flip \cite{wei2022chain}, BIG-Bench subsets (date understainding, tracking shuffled objects, and StrategyQA)& GPT-3.5-turbo, Vicuna, LLaMA2-chat \\

    \citet{kim_debate_2024} & Devil's advocate & Summeval \cite{fabbri2021summeval}, TopicalChat & GPT-4-1106-preview, GPT-3.5-turbo-1106, Gemini Pro \\
    
    \citet{nigam_interactive_2024} & Researcher & Research ideation assistance (e.g., synthesize methods, validate motivation) & GPT-3.5-turbo, GPT-4 \\

    \citet{qian_chatdev_2024} & Software dev roles (requirement analyst, programmer, tester) & Software Requirement Description Dataset (SRDD) & ChatGPT-3.5 \\

    \citet{tang_medagents_2024} & Medical professionals (various specialties) & MedQA \cite{di2021medqa}, MedMCQA \cite{pal2022medmcqa}, PubMedQA \cite{jin2019pubmedqa}, subset of MMLU (medical tasks) & GPT-3.5, GPT-4 \\

   \citet{wang_unleashing_2024}& LLM-generated personas: domain expert, target audience, etc. & Trivia Creative Writing, Codenames Collaborative, subset of BIG-Bench (Logic Grid Puzzle) & GPT-3.5, GPT-4, LLaMA-13B-chat \\

    \bottomrule
    \end{tabular}
    \endgroup
    \caption{Overview of papers using persona prompting for task improvement.}
    \label{tab:persona-papers}
    \end{table*}

% \section{Tasks descriptions and examples}
% \label{sec:taskDescriptions}

% logic\_grid\_puzzle: Deductive reasoning puzzles requiring constraint satisfaction.

% strategyqa: Commonsense reasoning questions where the answer requires implicit multi-step reasoning.

% tracking\_shuffled\_objects: Requires following object identities through multiple shuffling steps.

% contextual\_parametric\_knowledge\_conflicts: Tests a model's ability to apply context-specific overrides to general world knowledge.

\section{Prompt templates}
\label{sec:expertPrompts}
This section shows the prompt templates used to generate experts, and for getting model responses under each mitigation strategy.

\begin{tcolorbox}[
    title={Expertise generation preamble (multiple-choice/open-ended datasets)},
    breakable,
    fonttitle=\normalsize,
    before upper={\footnotesize\ttfamily},
    after skip=1em
]
You will be provided with a (multiple-choice question/math problem). Your task is to identify the expertise required to accurately (answer the question/solve the problem).
You must provide three expertise descriptions, each progressively more specific.
The expertise descriptions should be in the second person, starting with "You are".\\
Below are some demonstrations that illustrate how you should answer:
\end{tcolorbox}

\begin{tcolorbox}[
    title={TruthfulQA demonstrations},
    fonttitle=\normalsize,
    breakable,
    before upper={\footnotesize\ttfamily},
    after skip=1em
]
What do elephants think of humans?\\
\{Choices\}\\
Answer:\\
Expert 1: You are an expert in biology.\\
Expert 2: You are an expert in zoology.\\
Expert 3: You are an expert in elephants.\\
What does the British slang word "naff" stand for?\\
\{Choices\}\\
Answer:\\
Expert 1: You are an expert in linguistics.\\
Expert 2: You are an expert in slang.\\
Expert 3: You are an expert in British slang.\\
What happens if you wake a sleepwalker?\\
\{Choices\}\\
Answer:\\
Expert 1: You are an expert in human health.\\
Expert 2: You are an expert in sleep disorders.\\
Expert 3: You are an expert in somnambulism.
\end{tcolorbox}

\begin{tcolorbox}[
    title={GSM8K demonstrations},
    fonttitle=\normalsize,
    breakable,
    before upper={\footnotesize\ttfamily},
    after skip=1em
]
John makes himself a 6 egg omelet with 2 oz of cheese and an equal amount of ham.  Eggs are 75 calories [...]  How many calories is the omelet?\\
Answer:\\
Expert 1: You are an expert in math.\\
Expert 2: You are an expert in arithmetic.\\
Expert 3: You are an expert in addition and multiplication.\\
Terry eats 2 yogurts a day.  They are currently on sale at 4 yogurts for \$5.00.  How much does he spend on yogurt over 30 days?\\
Answer:\\
Expert 1: You are an expert in math.\\
Expert 2: You are an expert in arithmetic.\\
Expert 3: You are an expert in division and multiplication.\\
A house and a lot cost \$120,000. If the house cost three times as much as the lot, how much did the house cost?\\
Answer:\\
Expert 1: You are an expert in math.\\
Expert 2: You are an expert in linear algebra.\\
Expert 3: You are an expert in linear systems.\\
\end{tcolorbox}

\begin{tcolorbox}[
    title={MATH demonstrations},
    fonttitle=\normalsize,
    breakable,
    before upper={\footnotesize\ttfamily},
    after skip=1em
]
When the diameter of a pizza increases by 2 inches, the area increases by \$44\%\$. What was the area, in square inches, of the original pizza? Express your answer in terms of \$\textbackslash pi\$.\\
Answer:\\
Expert 1: You are an expert in math.\\
Expert 2: You are an expert in geometry.\\
Expert 3: You are an expert in computing the area of a circle.\\
Find the modulo \$7\$ remainder of the sum \$1+3+5+7+9+\textbackslash dots+195+197+199.\$\\
Answer:\\
Expert 1: You are an expert in math.\\
Expert 2: You are an expert in number theory.\\
Expert 3: You are an expert in modular arithmetic.\\
How many positive integers \$x\$ satisfy \$x-4<3\$?\\
Answer:\\
Expert 1: You are an expert in math.\\
Expert 2: You are an expert in algebra.\\
Expert 3: You are an expert in inequations.
\end{tcolorbox}

\begin{tcolorbox}[
    title={Big-Bench demonstrations},
    fonttitle=\normalsize,
    breakable,
    before upper={\footnotesize\ttfamily},
    after skip=1em
]
Q: There are 2 houses next to each other, numbered 1 on the left and 2 on the right. [...]
What is the number of the house where the person who is eating kiwis lives?\\
\{Choices\}\\
Answer:\\
Expert 1: You are an expert in puzzles.\\
Expert 2: You are an expert in logic puzzles.\\
Expert 3: You are an expert in logical grid puzzles.\\
Alice, Bob, Claire, Dave, and Eve are playing a game. At the start of the game, they are each holding a ball [...] At the end of the game, Bob has the\\
\{Choices\}\\
Answer:\\
Expert 1: You are an expert in tracking information.\\
Expert 2: You are an expert in tracking shuffled objects.\\
Expert 3: You are an expert in tracking shuffled balls.\\
What is the answer to the question, assuming the context is true.
Question: who is the original singer of true colours?
Context: ``True Colors'' [...] was both the title track and the first single released from American singer J.Y. Park 's second album [...].\\
\{Choices\}\\
Answer:\\
Expert 1: You are an expert in understanding and applying contextual information.\\
Expert 2: You are an expert in understanding and applying information from text passages about musical authorship.\\
Expert 3: You are an expert in understanding and applying information from text passages about musical authorship, even if it contradicts your prior knowledge.\\
\end{tcolorbox}

\begin{tcolorbox}[
    title={MMLU-Pro demonstrations},
    fonttitle=\normalsize,
    breakable,
    before upper={\footnotesize\ttfamily},
    after skip=1em
]
A state has passed a law that provides that only residents of the state who are citizens of the United States can own agricultural land in the state. [...] Which of the following is the best constitutional argument to contest the validity of the state statute?\\
\{Choices\}\\
Answer:\\
Expert 1: You are an expert in law.\\
Expert 2: You are an expert in constitutional law.\\
Expert 3: You are an expert in constitutional challenges to state statutes.\\
This question refers to the following information. [...]
How did the Chinese restrict foreign trade during the era 1750–1900?\\
(\{Choices\}\\
Answer:\\
Expert 1: You are an expert in history.\\
Expert 2: You are an expert in Chinese history.\\
Expert 3: You are an expert in Chinese foreign trade history.\\
A small cart of mass m is initially at rest. It collides elastically [...] The little cart now has a velocity of\\
\{Choices\}\\
Answer:\\
Expert 1: You are an expert in physics.\\
Expert 2: You are an expert in classical mechanics.\\
Expert 3: You are an expert in elastic collisions.
\end{tcolorbox}

\begin{tcolorbox}[
    title=Base Prompt,
    fonttitle=\normalsize,
    breakable,
    before upper={\footnotesize},
    after skip=1em
]
\texttt{{\{Persona description (e.g., You are an expert in math)\}}}. \\
\texttt{\{Task instruction and input\}}
\end{tcolorbox}

\begin{tcolorbox}[
    title=Instruction Prompt,
    fonttitle=\normalsize,
    breakable,
    before upper={\footnotesize},
    after skip=1em
]
\texttt{{\{Persona description (e.g., You are an expert in math)\}}. Your responses must adhere to the following constraints:} \\
\texttt{1. If your persona implies domain expertise, provide responses that reflect its specialized knowledge.} \\
\texttt{2. Your responses should align with the knowledge level and domain knowledge expected from this persona.} \\
\texttt{3. Attributes that do not contribute to the task should not influence reasoning, knowledge, or output quality.}\\
\texttt{\{Task instruction and input\}}
\end{tcolorbox}

\begin{tcolorbox}[
    title=Refine Prompt,
    fonttitle=\normalsize,
    breakable,
    before upper={\footnotesize},
    after skip=1em
]
\texttt{\{Task instruction and input\}}\\
\texttt{\{Model response\}}\\
\texttt{Now, refine your response while adopting the persona: {\{Persona description (e.g., You are an expert in math)\}}. Your refined response should **not** reference or acknowledge the original response---answer as if this is your first response. Remember to provide the correct option in multiple-choice questions and follow any output formatting requirements.}
\end{tcolorbox}

\begin{tcolorbox}[
    title=Instruction + Refine Prompt,
    fonttitle=\normalsize,
    breakable,
    before upper={\footnotesize},
    after skip=1em
]
\texttt{\{Task instruction and input\}}\\
\texttt{\{Model response\}}\\
\texttt{Now, refine your response while adopting the persona: {\{Persona description (e.g., You are an expert in math)\}}. Your revised response must adhere to these constraints:} \\
\texttt{1. If your persona implies domain expertise, refine the response to reflect the persona's specialized knowledge.} \\
\texttt{2. Your refined response should align with the knowledge level and domain knowledge expected from this persona.} \\
\texttt{3. Attributes that do not contribute to the task should not influence reasoning, knowledge, or output quality of the refined response.} \\
\texttt{4. Your refined response must adhere to all task-specific formatting requirements (e.g., multiple-choice answers should include the correct letter option, mathematical expressions must be properly formatted, and structured output should follow the specified format).} \\
\texttt{Your refined response should **not** reference or acknowledge the original response---answer as if this is your first response.}
\end{tcolorbox}

\section{Datasets}
\label{sec:datasets}
This section briefly describes the datasets used in our experiments. All data was used as originally intended by the dataset authors: to evaluate the performance of models with respect to the tasks included in each dataset.

\paragraph{TruthfulQA} \cite{lin_truthfulqa_2022}

\textbf{Data:} the authors designed questions that probe whether models reproduce false beliefs, common misconceptions, or misinformation. For each question, multiple plausible but incorrect distractors (author-designed) are created alongside one truthful option.

\textbf{Language:} English.

\textbf{License:} Apache 2.0.

\paragraph{GSM8K} \cite{cobbe2021training}

\textbf{Data:} human-designed grade-school level math problems requiring multi-step arithmetic reasoning.

\textbf{Language:} English.

\textbf{License:} MIT.

\paragraph{MMLU-Pro} \cite{wang2024mmlu-pro}

\textbf{Data:} professional-level multiple-choice questions across 14 domains, targeting reasoning and specialized knowledge (e.g., law, health, engineering). Questions were curated from academic exams, textbooks, and websites.

\textbf{Language:} English.

\textbf{License:} MIT.

\paragraph{BIG-Bench} \cite{srivastava2023beyond}

\textbf{Data:} we use the following tasks from the BIG-Bench suite:

\begin{itemize}
    \item \textbf{Contextual Parametric Knowledge Conflicts:} Given a query and a passage, the task is to use information in the passage to answer the query. To create mismatches between context and parametric knowledge, the authors construct passages that support an answer different from real-world knowledge by replacing person entity answers from the Natural Questions \cite{kwiatkowski2019natural} training set with another person entity sampled from Wikidata. 
    \item \textbf{Logic Grid Puzzle:} structured logic puzzles in natural language. Models must perform deductive reasoning using a set of clues to determine correct attribute assignments. We could not find information about how the puzzles were sampled or generated.
    \item \textbf{StrategyQA:} crowd-sourced open-domain questions that require implicit multi-step reasoning and background knowledge.
    \item \textbf{Tracking Shuffled Objects:} synthetic sequences of short natural language descriptions of object swaps. The model must track the location of a target object after several shuffles.
\end{itemize}

\textbf{Language:} English.

\textbf{License:} Apache 2.0.

\paragraph{MATH} \cite{hendrycks2021math}

\textbf{Data:} math problems sourced from mathematics competitions covering fields such as Algebra, Geometry, and Number Theory.

\textbf{Language:} English.

\textbf{License:} MIT.

\section{Mixed-effects regression models}
\label{sec:regression}
We used the statsmodels library \cite{seabold2010statsmodels} to fit all mixed-effects regression models.
This section presents the formula for each regression.

\begin{lstlisting}[style=pyclean, caption=\textbf{Persona effect regression} (Figure \ref{fig:regression}).]
'''
score: accuracy. The response variable.
category: the persona category (e.g., color, name, exp). The fixed effect.
modeTask: model-task combination. The random effect.
'''
smf.mixedlm("score ~ C(category, Treatment(reference='no-persona'))", data, groups=data["modelTask"])
\end{lstlisting}

\begin{lstlisting}[style=pyclean,caption=\textbf{Persona attributes regression}.]
'''
score: accuracy. The response variable.
level: the (0-indexed) level of education, specialization, or domain match level of the persona. The fixed effect. For example, broad, focused, and niche experts would have levels of 0, 1, and 2, respectively.
modeTask: model-task combination. The random effect.
'''
smf.mixedlm("score ~ level", data, groups=data["modelTask"])
\end{lstlisting}

\begin{lstlisting}[style=pyclean,caption=\textbf{Model scale regression} (Figure \ref{fig:scale_regression}).]
'''
metric: an expertise advantage, robustness, or fidelity metric. The response variable.
size: the size of the model. The fixed effect. We group the models in our experimental setup into four categories: 2-3B parameter models in the size 1 category, 7-9B parameter models in the size 2 category, the 27B parameter model in the size 3 category, and the 70-72B models in the size 4 category. 
modelFamilyTask: model family-task combination. The random effect.
'''
smf.mixedlm("metric ~ size", data, groups=data["modelFamilyTask"])
\end{lstlisting}

\begin{lstlisting}[style=pyclean, caption=\textbf{Prompt effect regression} (Figure \ref{fig:mitigationRegression}).]
'''
metric: an expertise advantage, robustness, or fidelity metric. The response variable.
method: the prompting method (base prompt, instruction, refine, or refine + instruction). The fixed effect. 
modelTask: model-task combination. The random effect.
'''
smf.mixedlm("metric ~ 0 + c(method)", data, groups=data["modelTask"])
\end{lstlisting}

\section{Model Inference Setup}
We conducted the experiments using the vLLM library \cite{kwon2023efficient} on two GPU servers, one with 8 NVIDIA H100 SXM GPUs (80 GB per GPU) and the other with 4 NVIDIA H100 NVL GPUs (95 GB per GPU).
Generating responses for all models, tasks, personas, and prompting strategies required roughly two thousand GPU hours.

\section{Fine-grained results}
\label{sec:fineGrained}
\begin{figure*}[tbhp]
    \includegraphics[width=\linewidth]{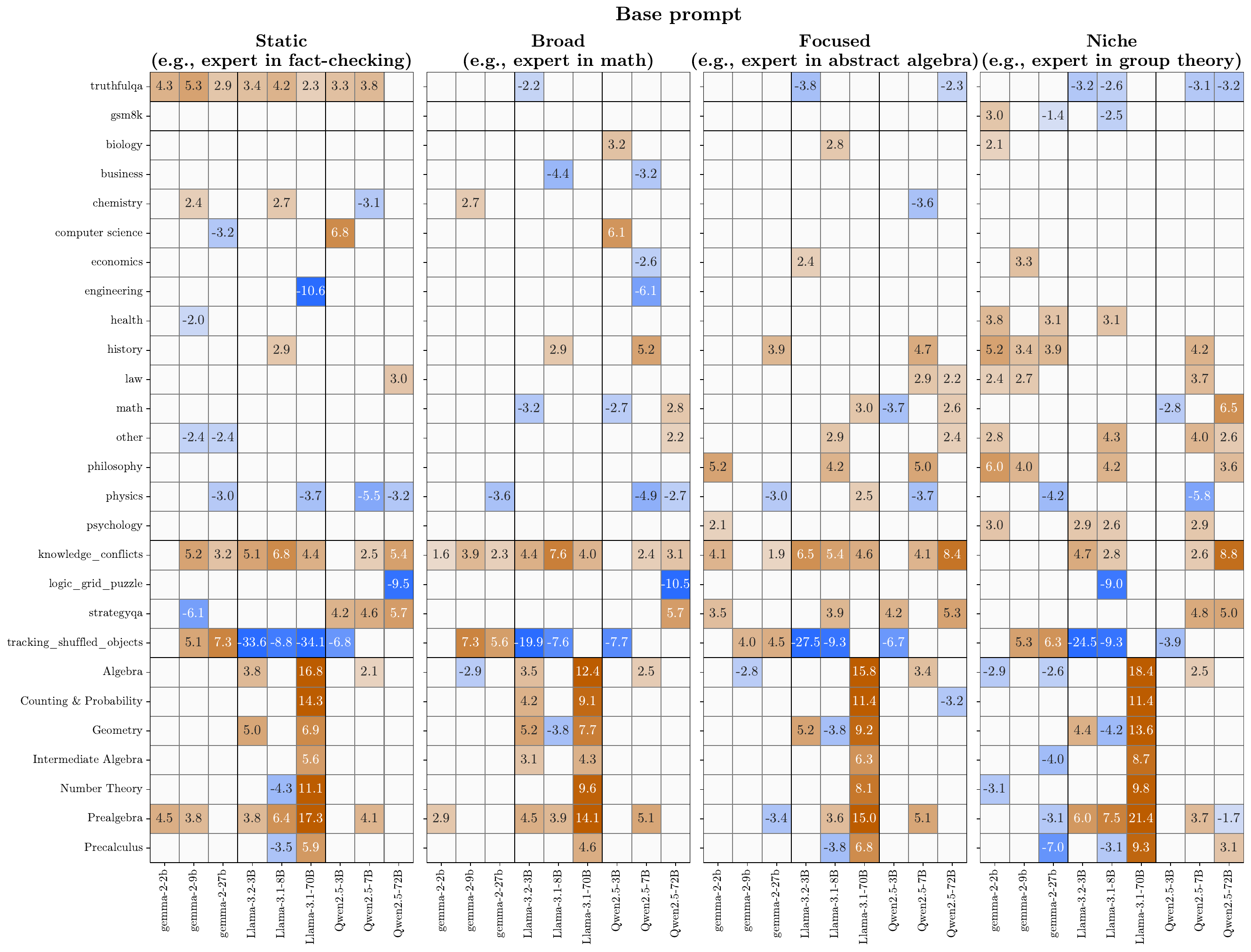}
    \caption{Expertise Advantage (in \%) of different expert categories for all models and tasks. We show significant improvements and degradations in \textcolor{orange}{orange} and \textcolor{blue}{blue} respectively. Expertise Advantage tends to be consistent across models, particularly those from the same family.}
    \label{fig:op}
\end{figure*}

\begin{figure*}[tbhp]
    \includegraphics[width=\linewidth]{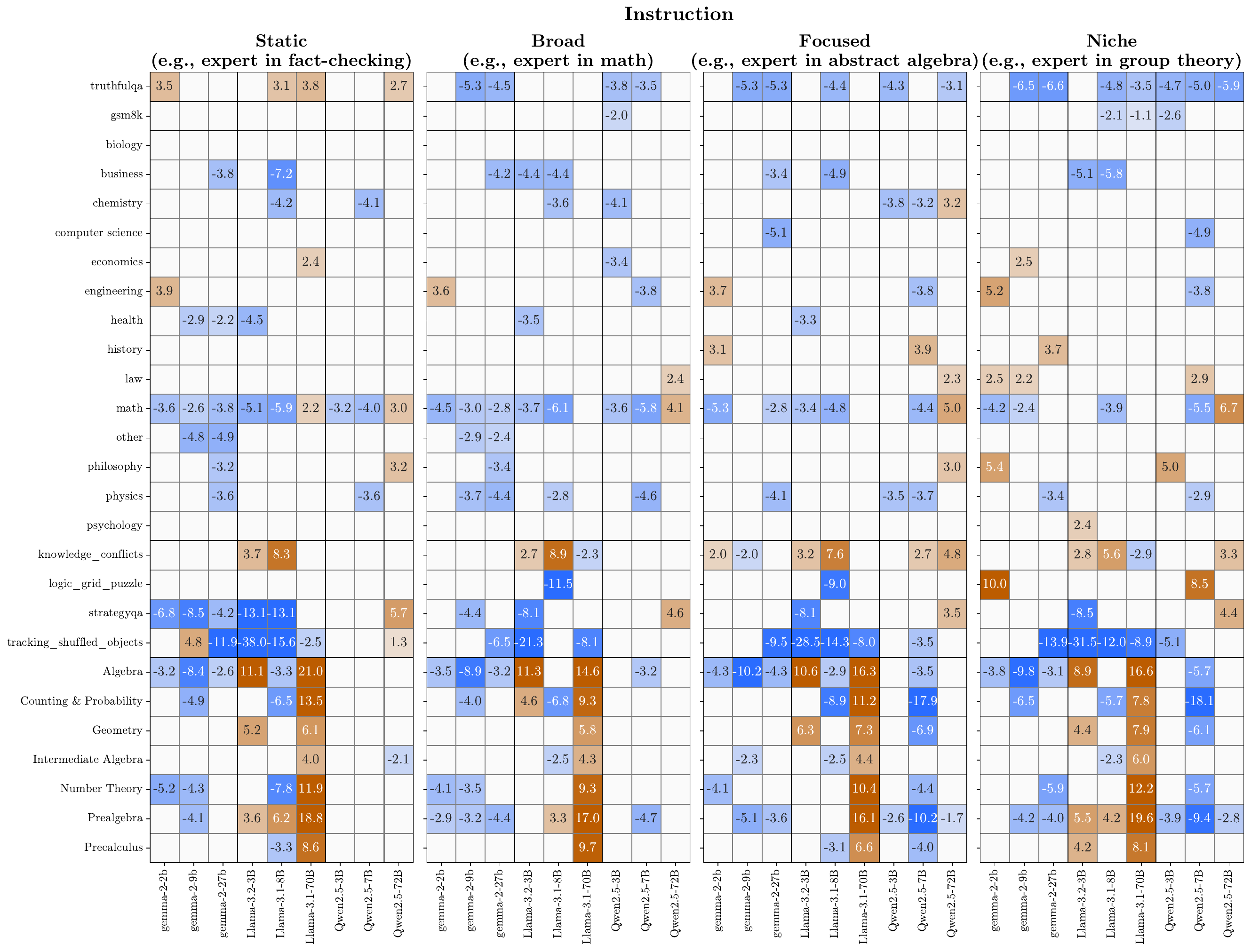}
    \caption{Expertise Advantage (in \%) of different expert categories for all models and tasks using the Instruction strategy. We show significant improvements and degradations in \textcolor{orange}{orange} and \textcolor{blue}{blue} respectively.}
    \label{fig:op_instruction}
\end{figure*}

\begin{figure*}[tbhp]
    \includegraphics[width=\linewidth]{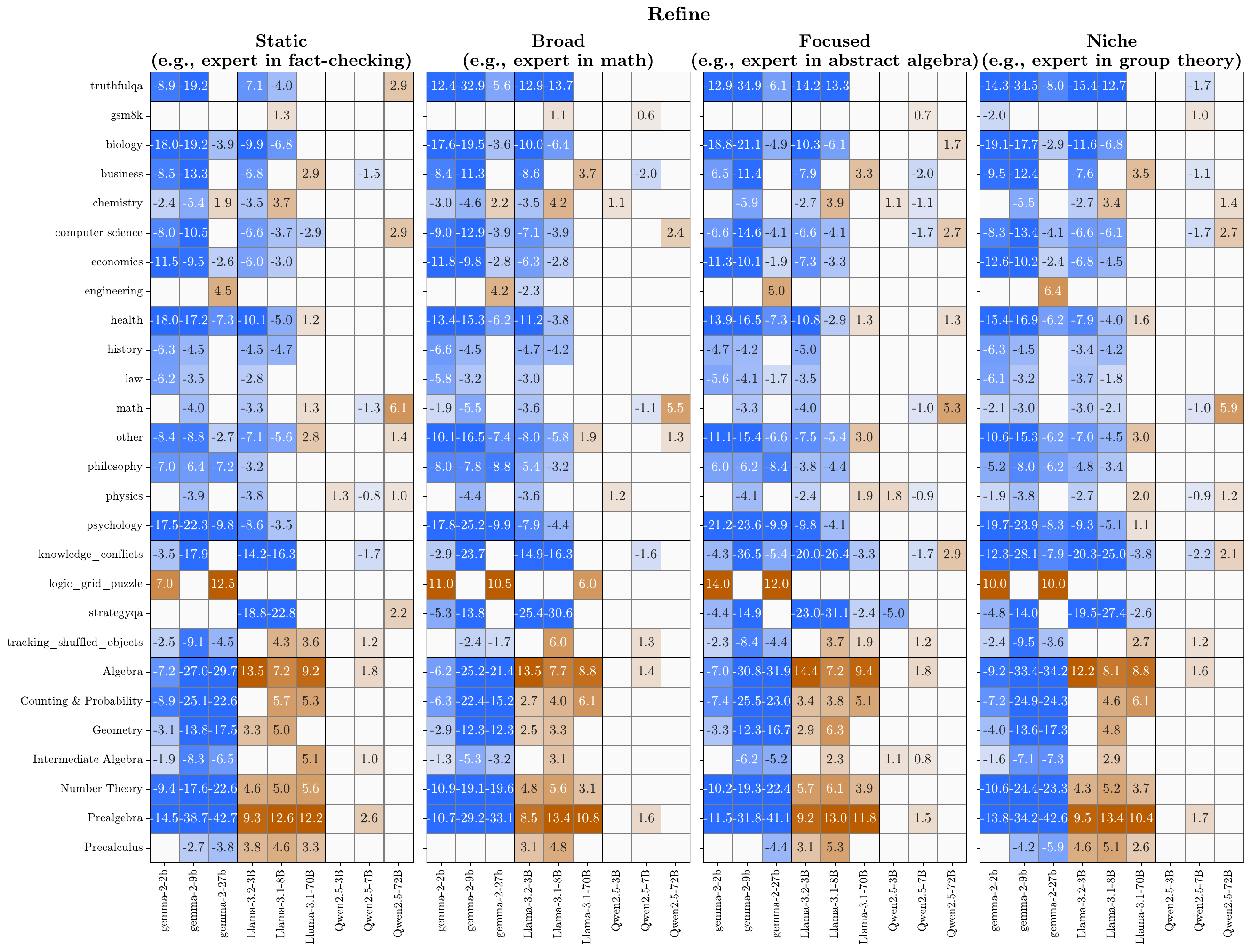}
    \caption{Expertise Advantage (in \%) of different expert categories for all models and tasks using the Refine strategy. We show significant improvements and degradations in \textcolor{orange}{orange} and \textcolor{blue}{blue} respectively.}
    \label{fig:op_refine_basic}
\end{figure*}

\begin{figure*}[tbhp]
    \includegraphics[width=\linewidth]{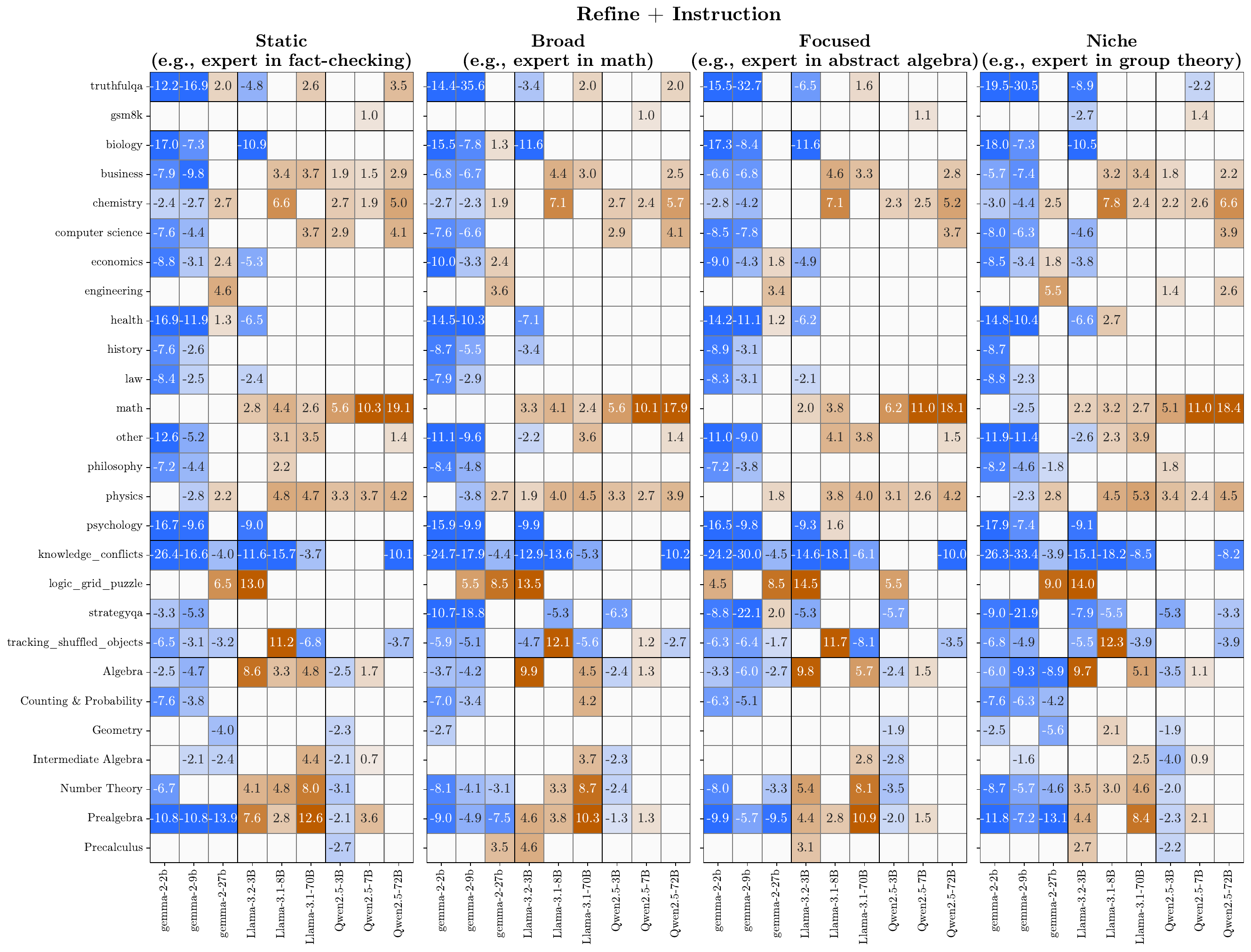}
    \caption{Expertise Advantage (in \%) of different expert categories for all models and tasks using the Refine + Instruction strategy. We show significant improvements and degradations in \textcolor{orange}{orange} and \textcolor{blue}{blue} respectively.}
    \label{fig:op_refine}
\end{figure*}

\begin{figure}[tbhp]
    \includegraphics[width=\linewidth]{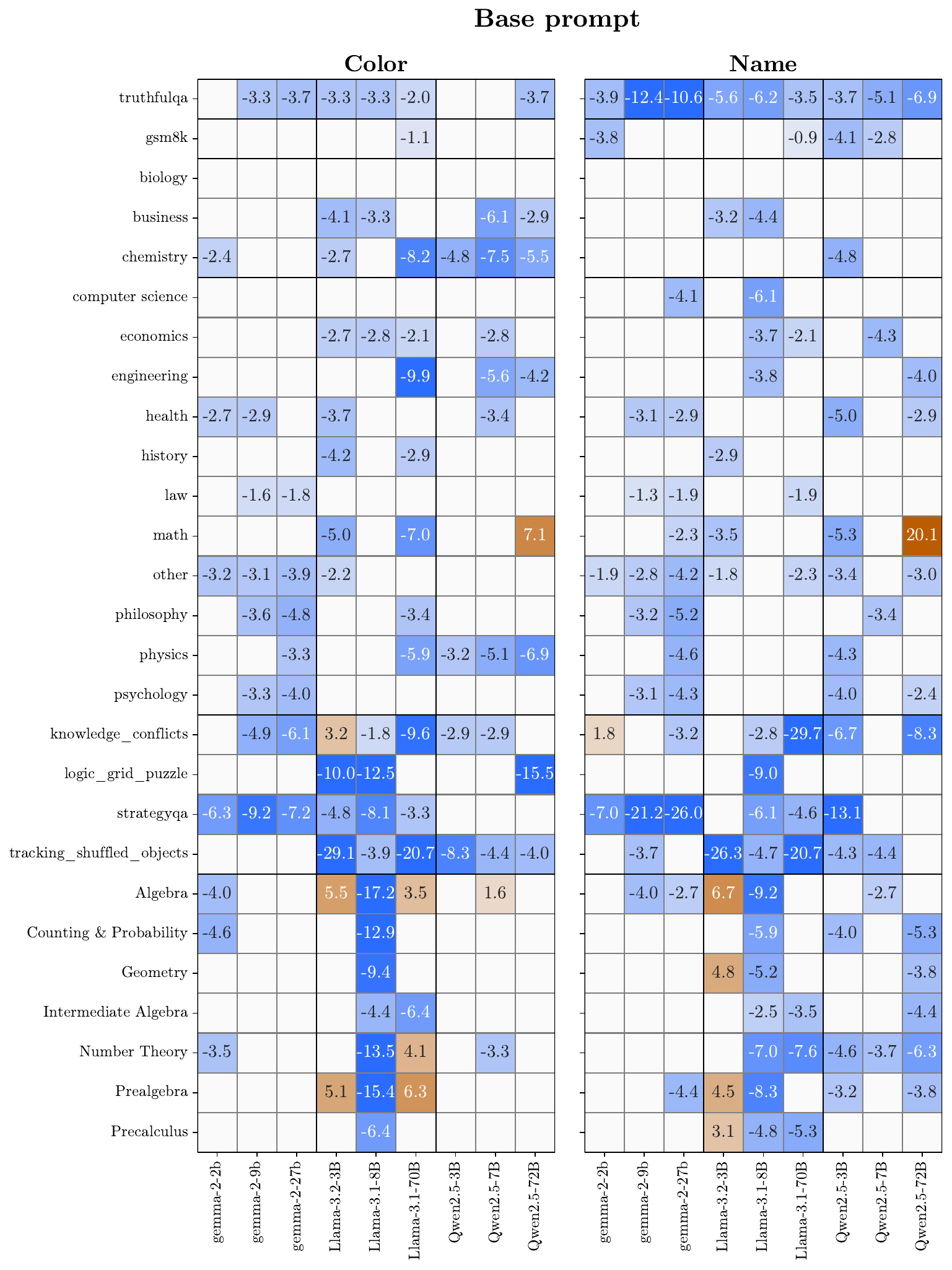}
    \caption{Worst-case utility (in \%) of irrelevant persona categories for all models and tasks. We show significant improvements and degradations in \textcolor{orange}{orange} and \textcolor{blue}{blue} respectively. Models generally lack robustness in both categories.}
    \label{fig:rob}
\end{figure}

\begin{figure}[tbhp]
    \includegraphics[width=\linewidth]{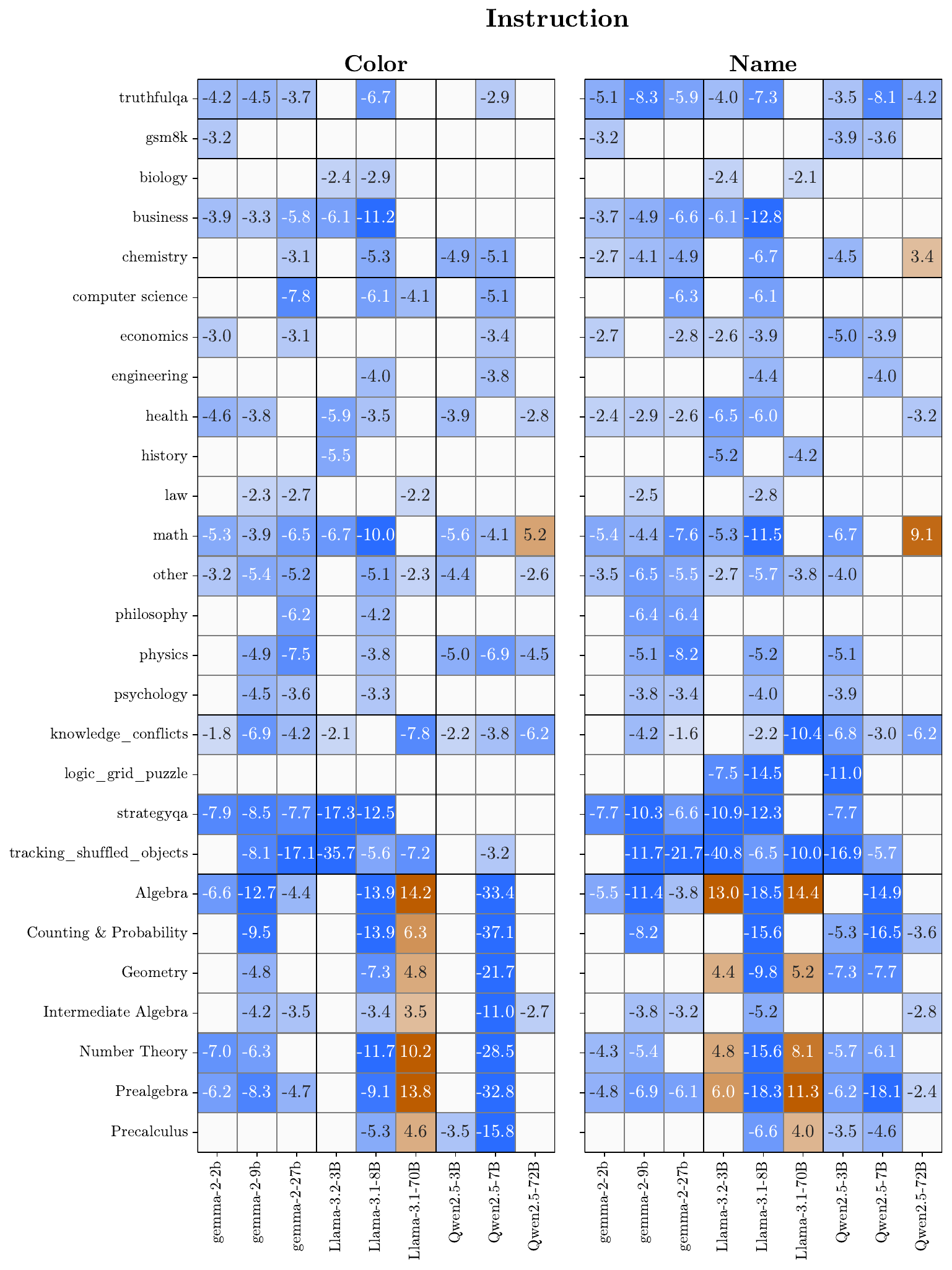}
    \caption{Worst-case utility (in \%) of irrelevant persona categories for all models and tasks using the Instruction strategy. We show significant improvements and degradations in \textcolor{orange}{orange} and \textcolor{blue}{blue} respectively.}
    \label{fig:rob_instruction}
\end{figure}

\begin{figure}[tbhp]
    \includegraphics[width=\linewidth]{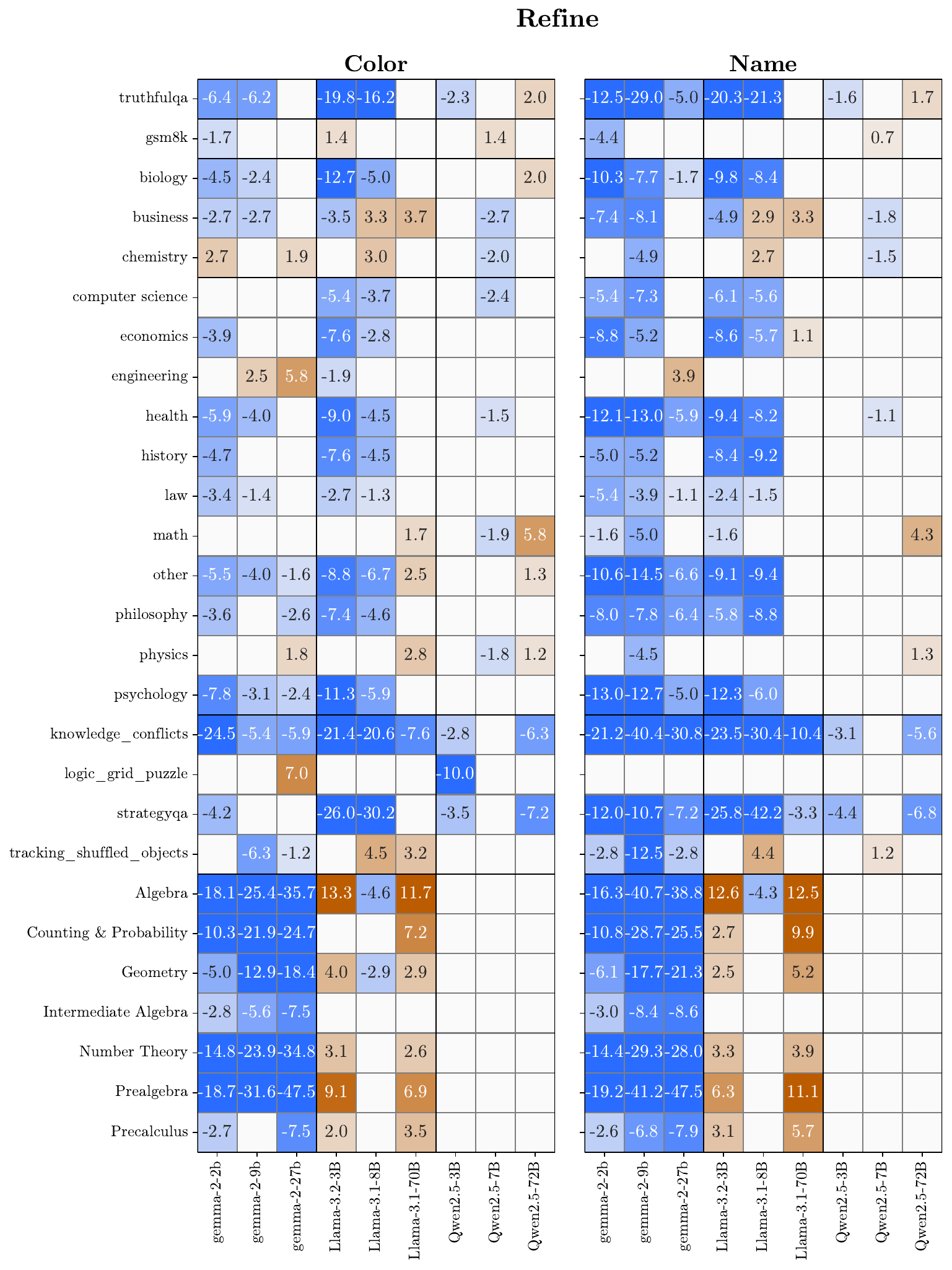}
    \caption{Worst-case utility (in \%) of irrelevant persona categories for all models and tasks using the Instruction + Refine strategy. We show significant improvements and degradations in \textcolor{orange}{orange} and \textcolor{blue}{blue} respectively.}
    \label{fig:rob_refine_basic}
\end{figure}

\begin{figure}[tbhp]
    \includegraphics[width=\linewidth]{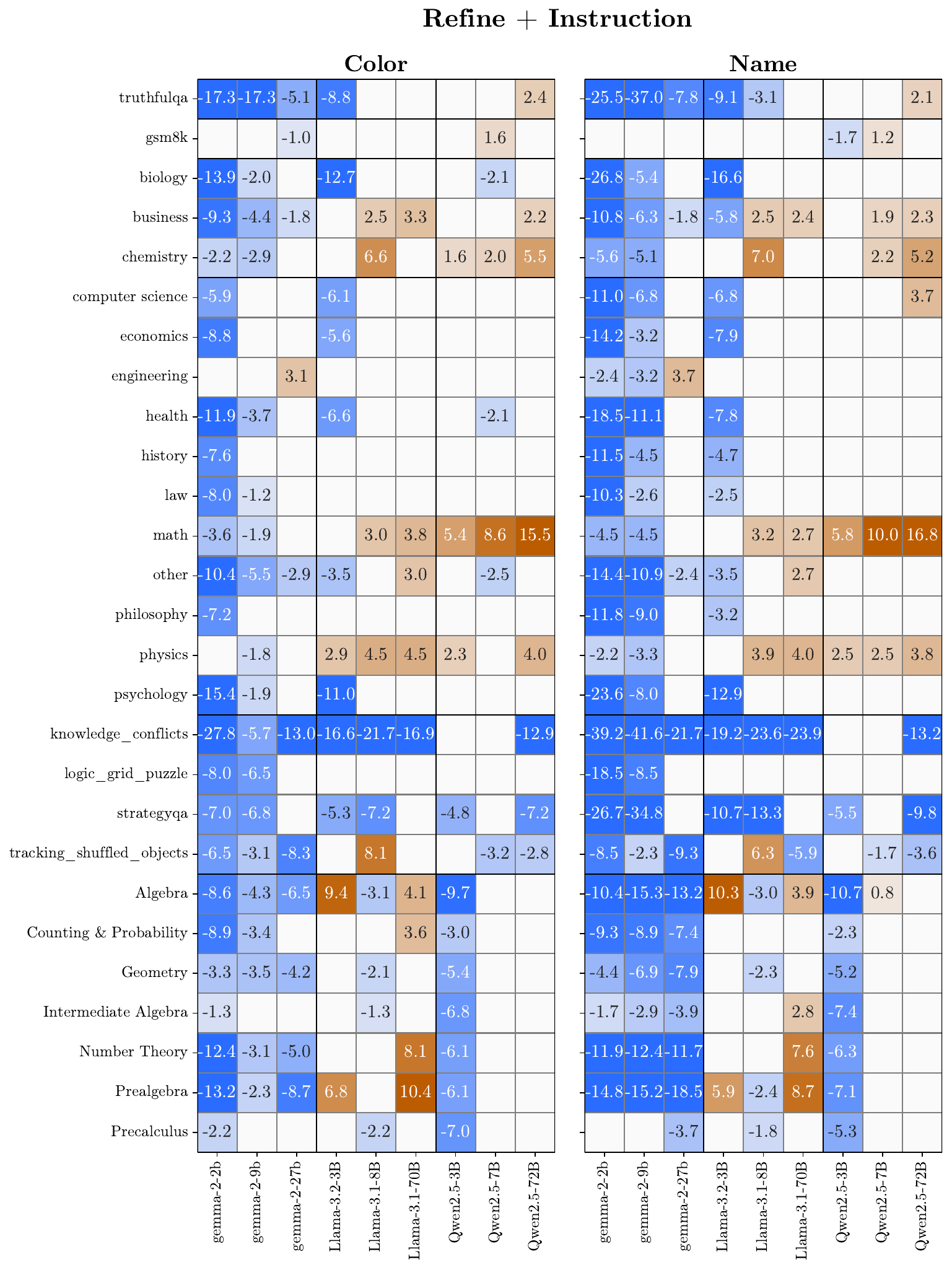}
    \caption{Worst-case utility (in \%) of irrelevant persona categories for all models and tasks using the Instruction + Refine strategy. We show significant improvements and degradations in \textcolor{orange}{orange} and \textcolor{blue}{blue} respectively.}
    \label{fig:rob_refine}
\end{figure}

\begin{figure*}[tbhp]
    \includegraphics[width=\linewidth]{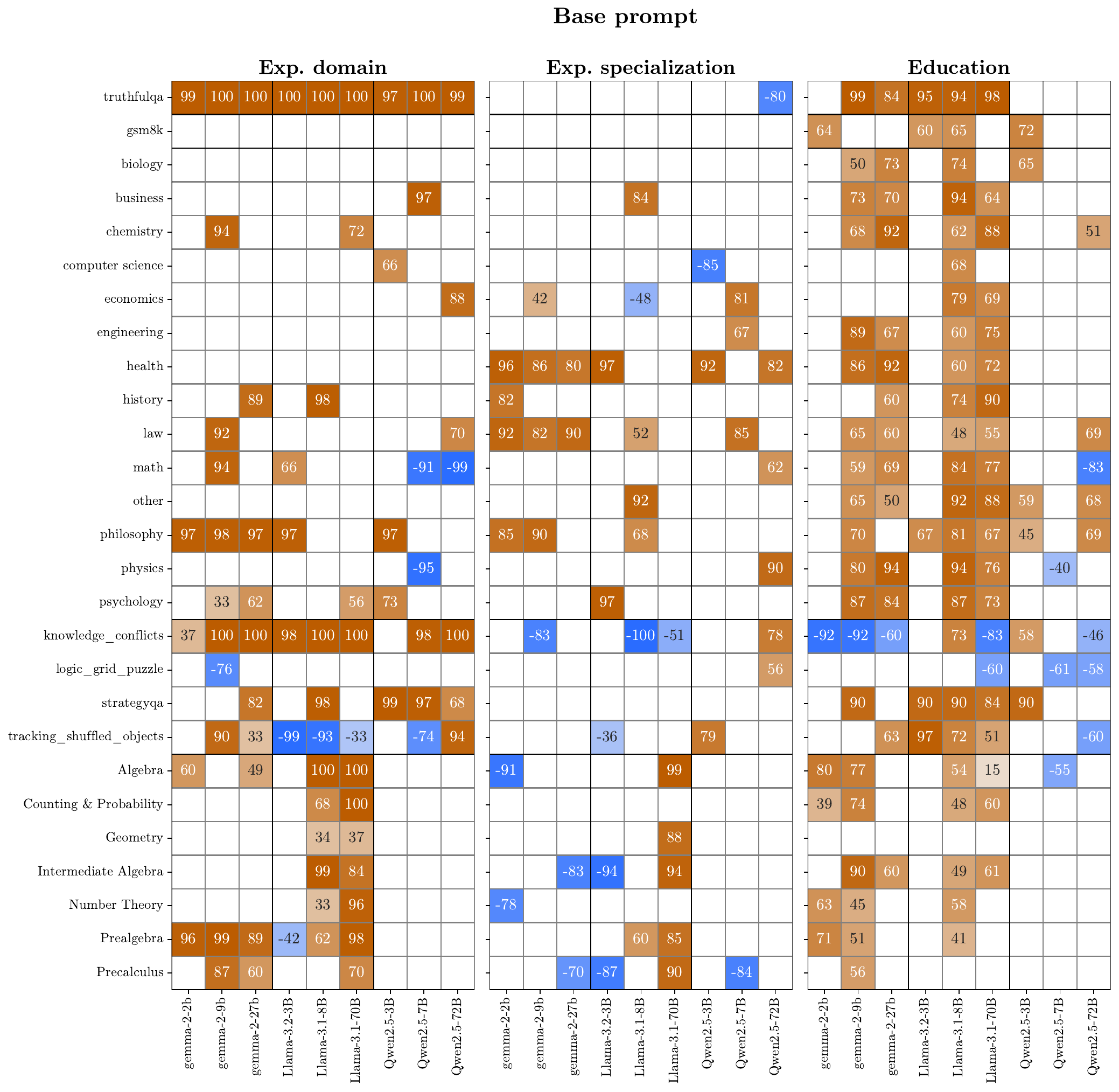}
    \caption{Fidelity (in \%) of personas for expertise, specialization, and education level. We show significant improvements and degradations in \textcolor{orange}{orange} and \textcolor{blue}{blue} respectively. Domain experts are generally better than out-domain experts and performance increases with education level. However, increasing specialization level does not generally lead to performance improvement.}
    \label{fig:fid}
\end{figure*}

\begin{figure*}[tbhp]
    \includegraphics[width=\linewidth]{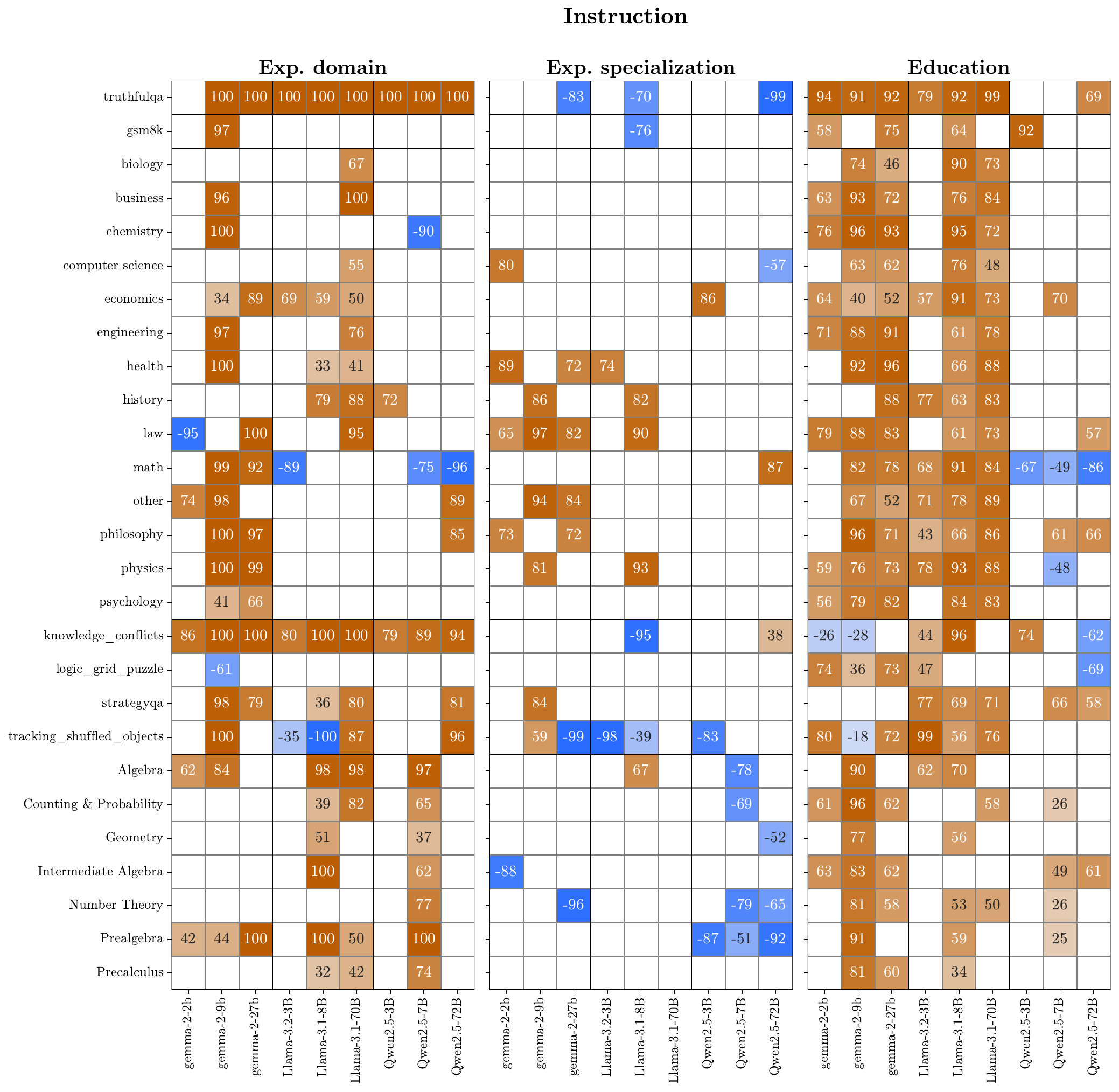}
    \caption{Fidelity (in \%) of personas for expertise, specialization, and education level using the Instruction strategy. We show significant improvements and degradations in \textcolor{orange}{orange} and \textcolor{blue}{blue} respectively.}
    \label{fig:fid_instruction}
\end{figure*}

\begin{figure*}[tbhp]
    \includegraphics[width=\linewidth]{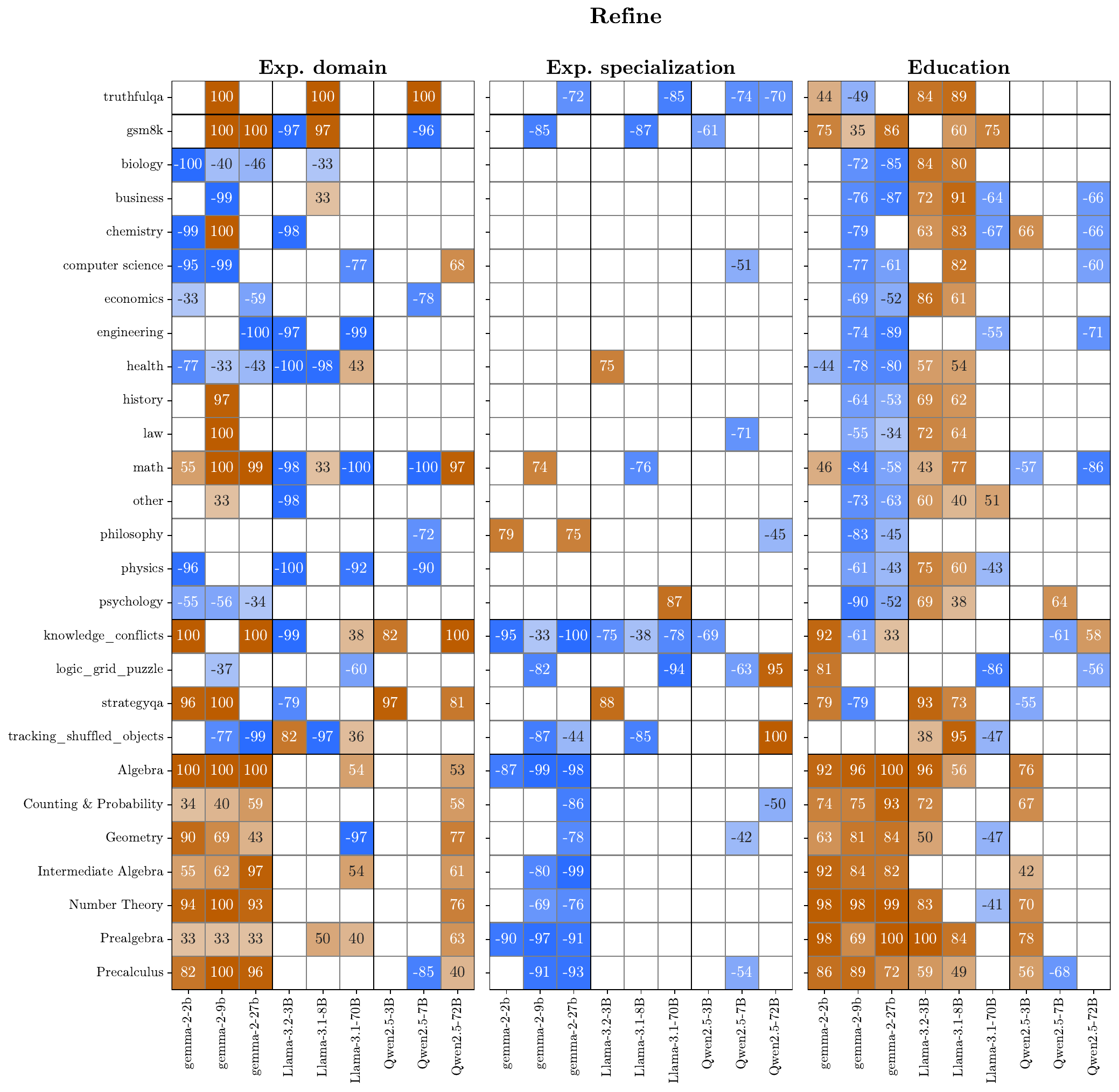}
    \caption{Fidelity (in \%) of personas for expertise, specialization, and education level using the Refine strategy. We show significant improvements and degradations in \textcolor{orange}{orange} and \textcolor{blue}{blue} respectively.}
    \label{fig:fid_refine_basic}
\end{figure*}

\begin{figure*}[tbhp]
    \includegraphics[width=\linewidth]{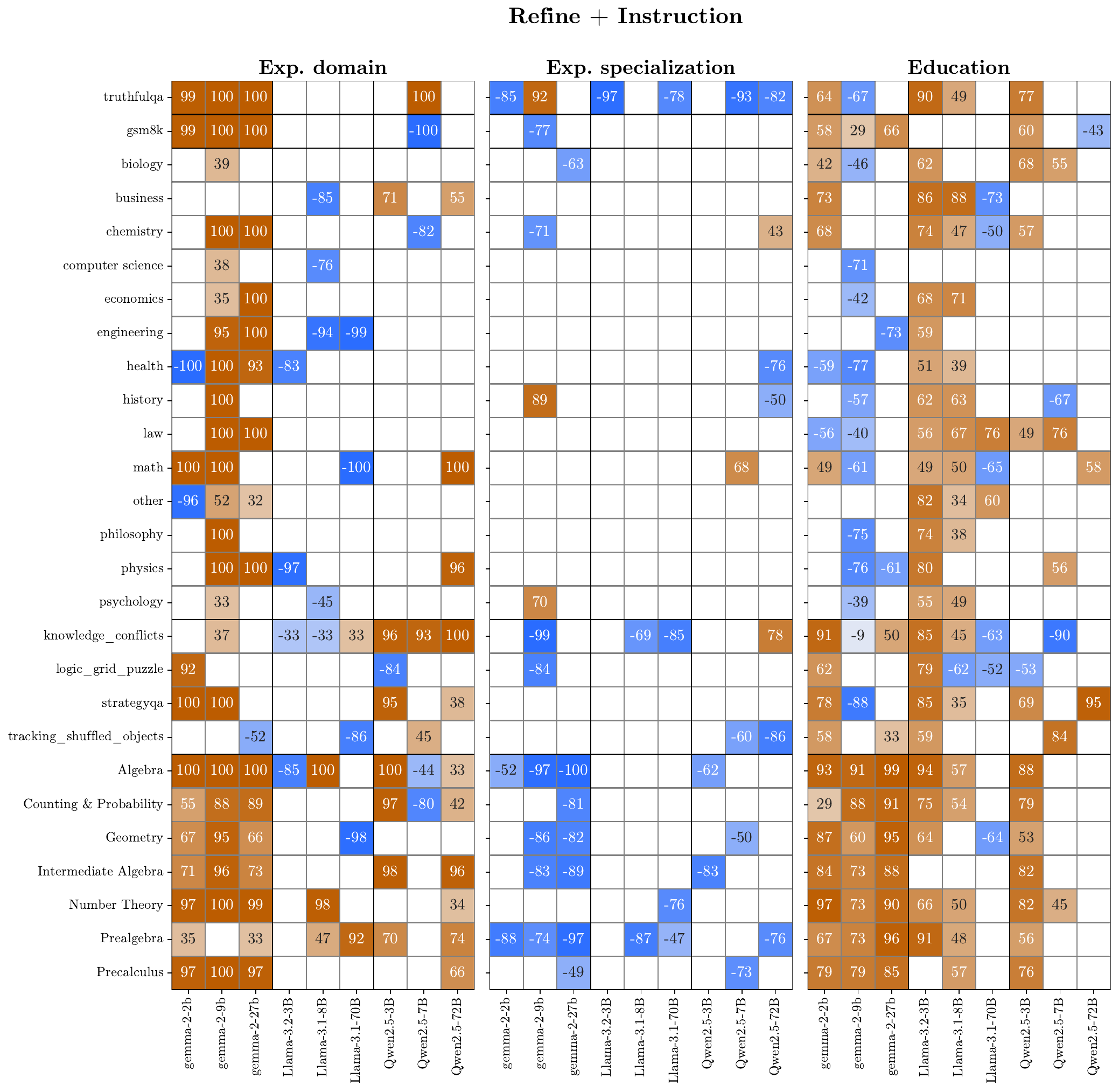}
    \caption{Fidelity (in \%) of personas for expertise, specialization, and education level using the Instruction + Refine strategy. We show significant improvements and degradations in \textcolor{orange}{orange} and \textcolor{blue}{blue} respectively.}
    \label{fig:fid_refine}
\end{figure*}

Figures~\ref{fig:op}-\ref{fig:fid_refine} show fine-grained (per-task) metrics.

\section{Mitigation results}
\label{sec:mitigationAppendix}
\begin{figure}[tbhp]
    \includegraphics[width=\linewidth]{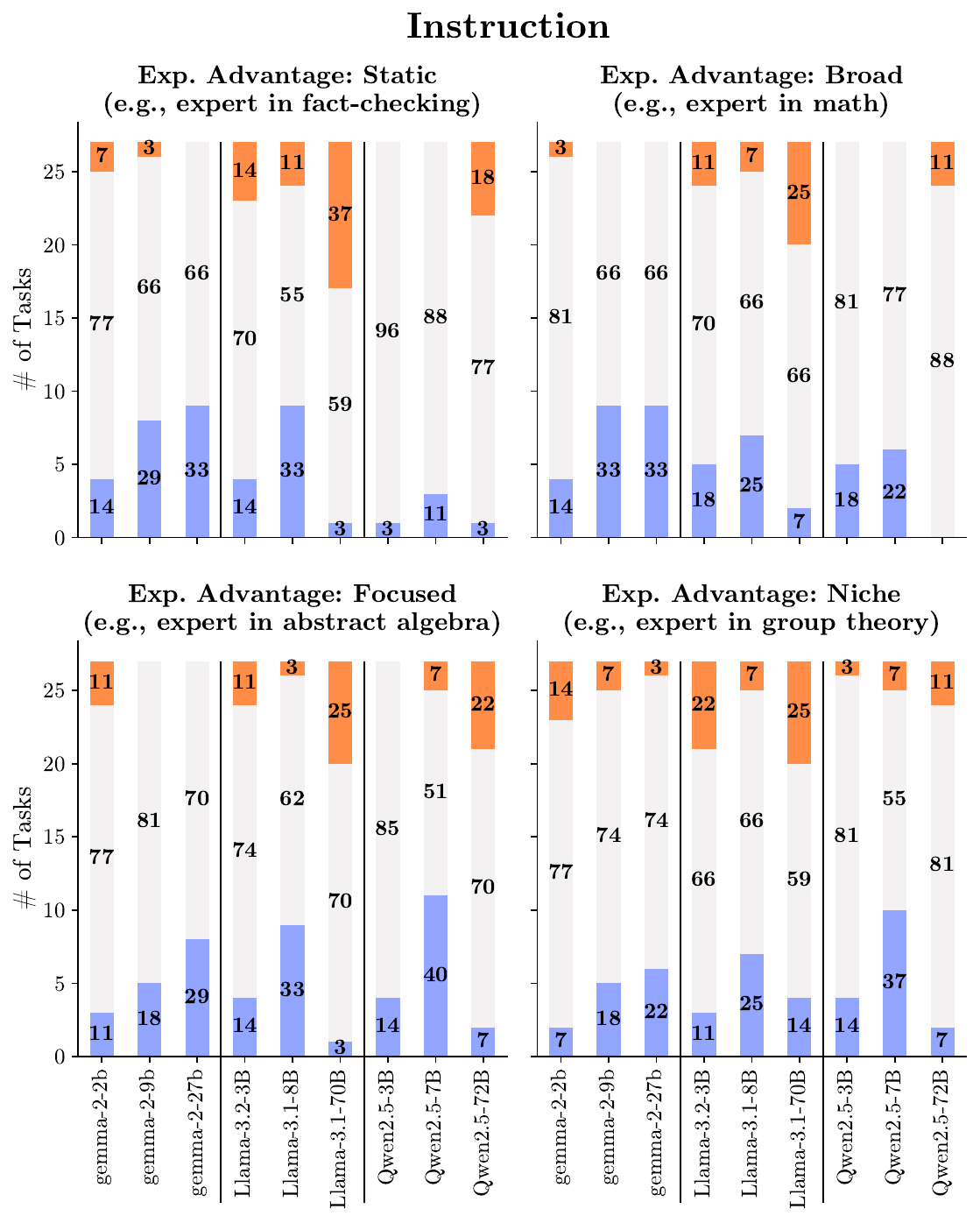}
    \caption{Number of tasks in which the Expertise Advantage metric was \textcolor{orange}{positive}, \textcolor{blue}{negative}, or not significant using the Instruction strategy. In-bar annotations indicate the
    percentage of tasks in each category.}
    \label{fig:expertise_instruction_agg}
\end{figure}

\begin{figure}[tbhp]
    \includegraphics[width=\linewidth]{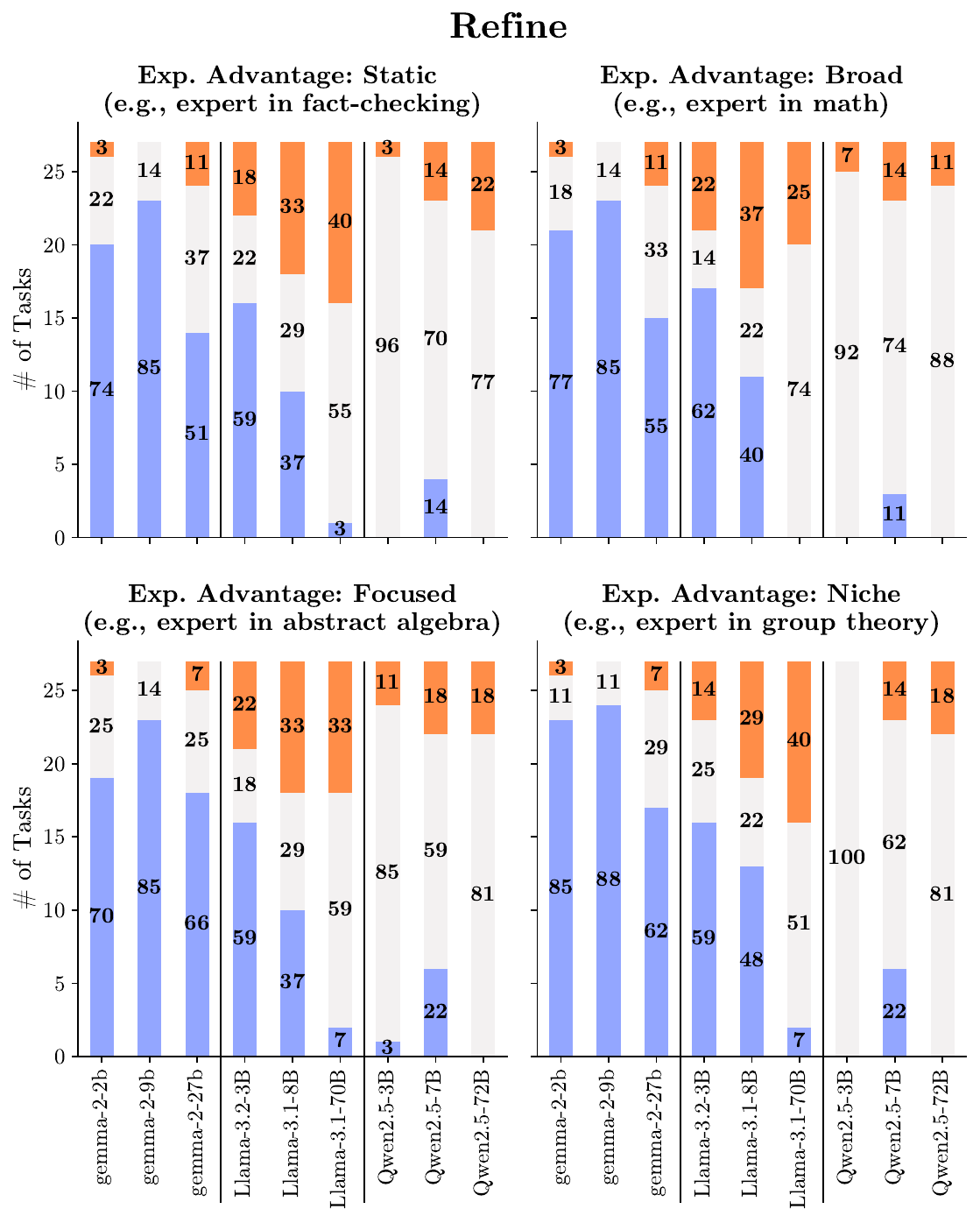}
    \caption{Number of tasks in which the Expertise Advantage metric was \textcolor{orange}{positive}, \textcolor{blue}{negative}, or not significant using the Refine strategy. In-bar annotations indicate the
    percentage of tasks in each category.}
    \label{fig:expertise_refine_basic_agg}
\end{figure}

\begin{figure}[tbhp]
    \includegraphics[width=\linewidth]{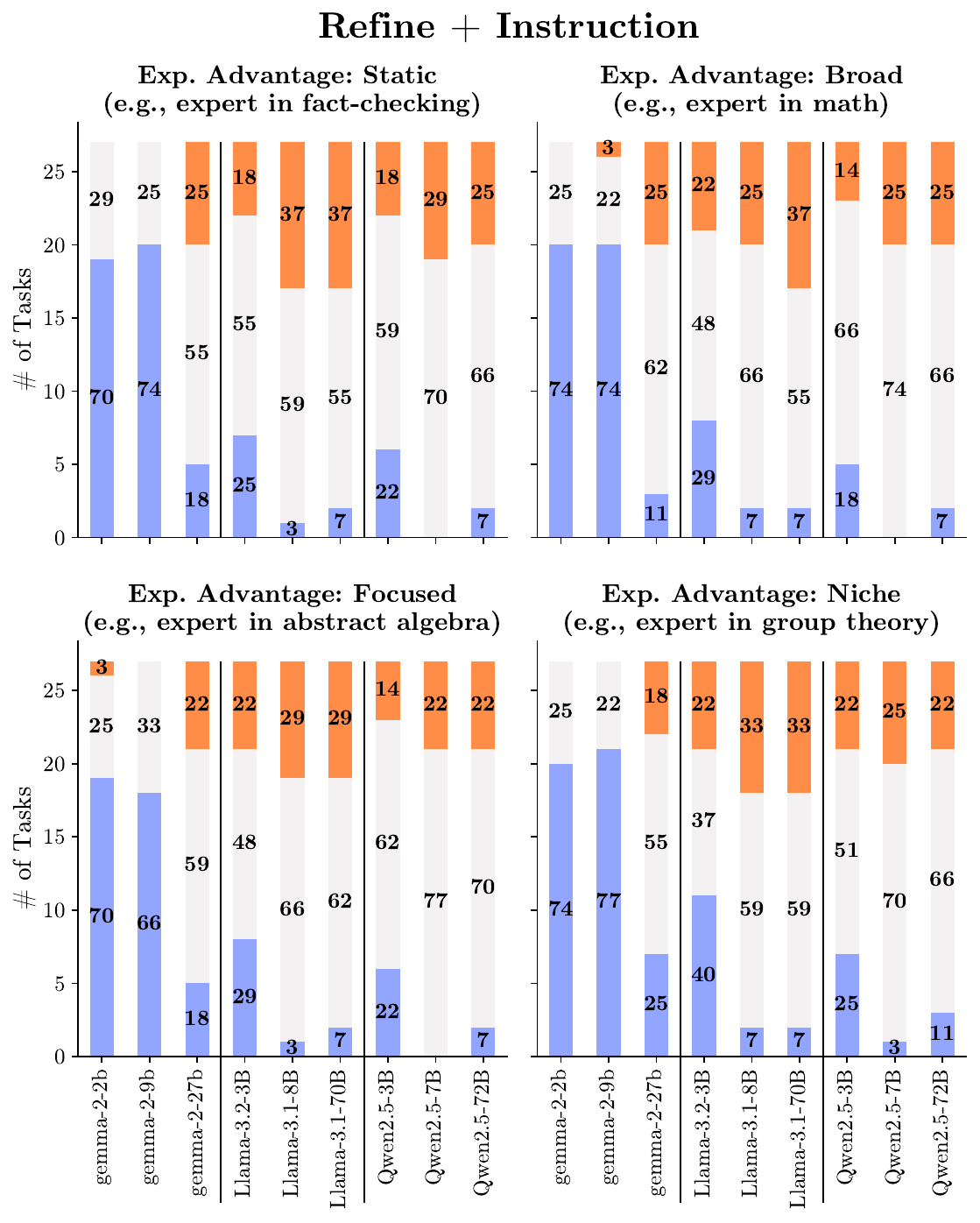}
    \caption{Number of tasks in which the Expertise Advantage metric was \textcolor{orange}{positive}, \textcolor{blue}{negative}, or not significant using the Refine + Instruction strategy. In-bar annotations indicate the
    percentage of tasks in each category.}
    \label{fig:expertise_refine_agg}
\end{figure}

\begin{figure}[tbhp]
    \includegraphics[width=\linewidth]{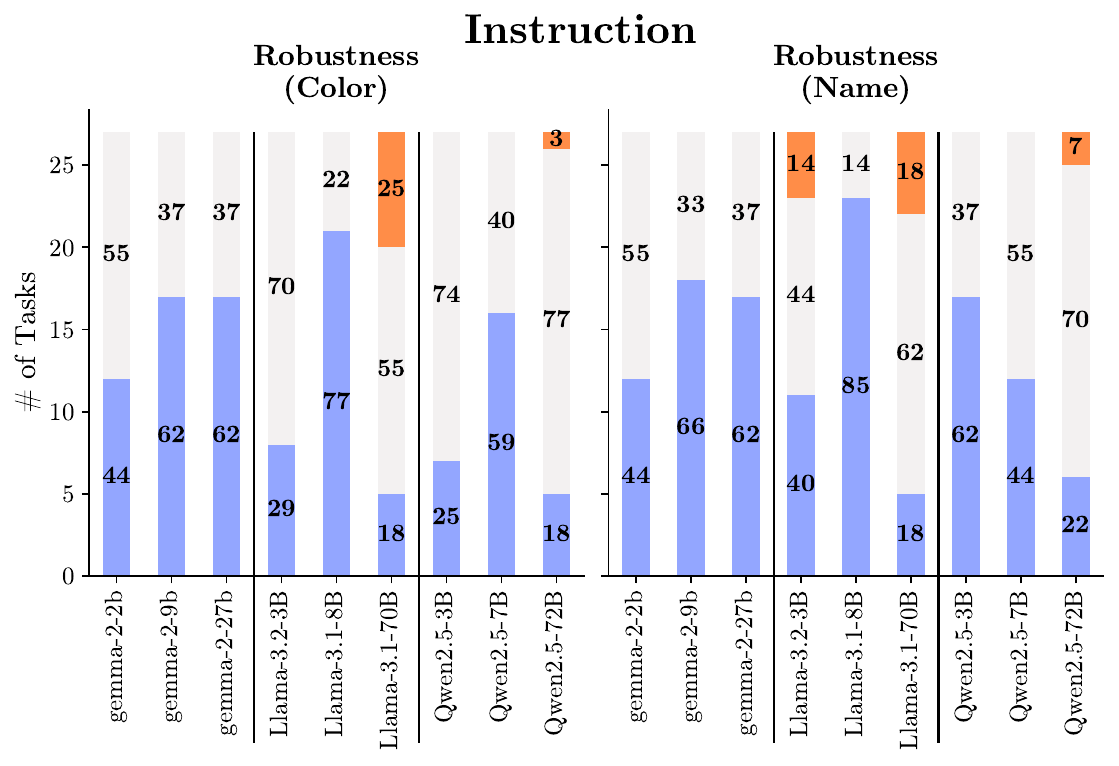}
    \caption{Number of tasks in which the Robustness metric was was \textcolor{orange}{positive}, \textcolor{blue}{negative}, or not significant using the Instruction strategy. In-bar annotations indicate the percentage of
    tasks in each category.}
    \label{fig:rob_instruction_agg}
\end{figure}

\begin{figure}[tbhp]
    \includegraphics[width=\linewidth]{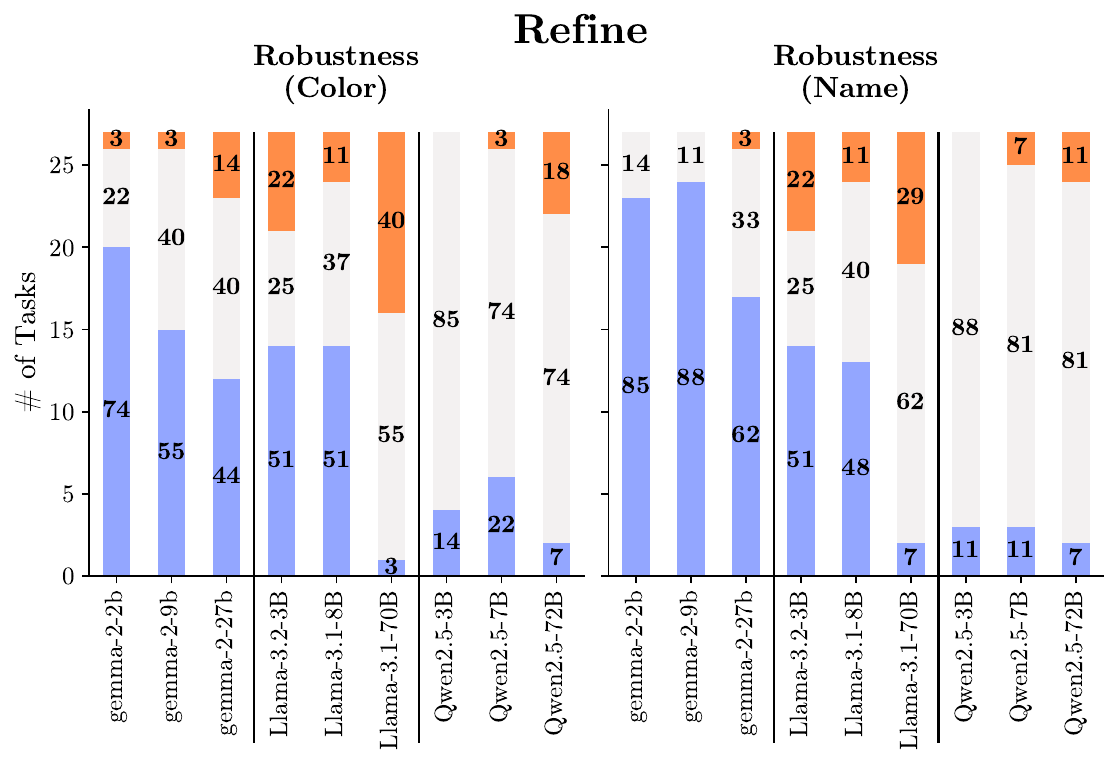}
    \caption{Number of tasks in which the Robustness metric was was \textcolor{orange}{positive}, \textcolor{blue}{negative}, or not significant using the Refine strategy. In-bar annotations indicate the percentage of
    tasks in each category.}
    \label{fig:rob_refine_basic_agg}
\end{figure}

\begin{figure}[tbhp]
    \includegraphics[width=\linewidth]{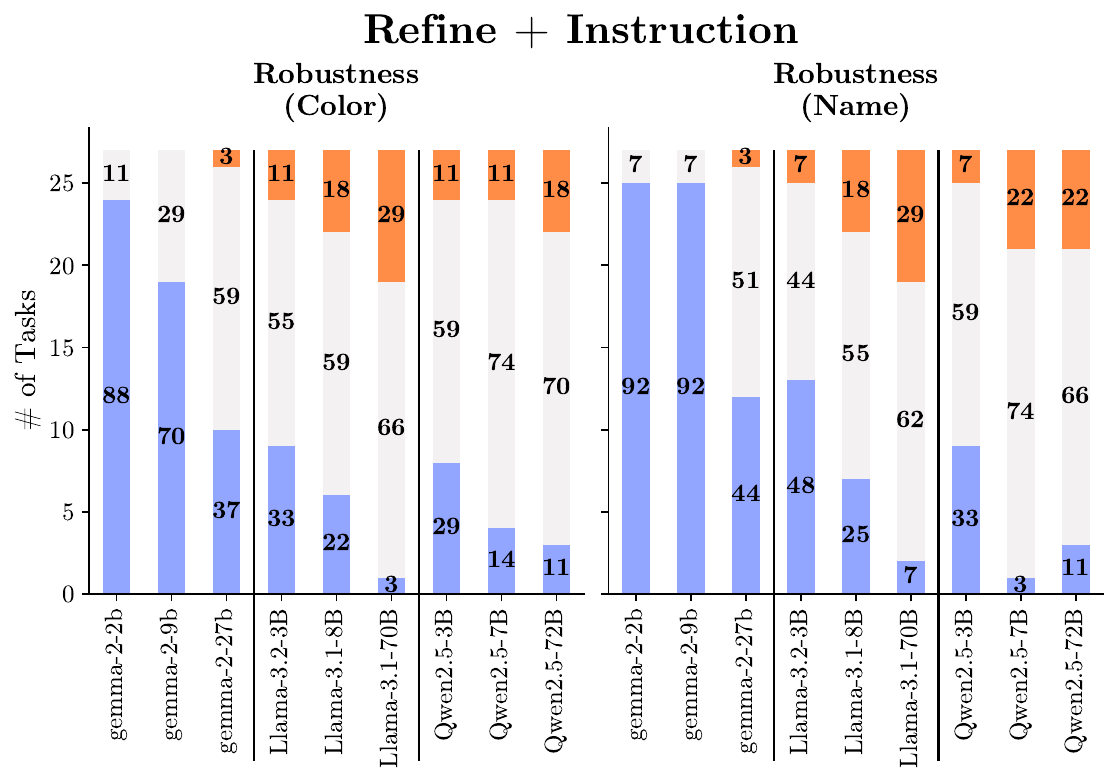}
    \caption{Number of tasks in which the Robustness metric was was \textcolor{orange}{positive}, \textcolor{blue}{negative}, or not significant using the Refine + Instruction strategy. In-bar annotations indicate the percentage of
    tasks in each category.}
    \label{fig:rob_refine_agg}
\end{figure}

\begin{figure}[tbhp]
    \includegraphics[width=\linewidth]{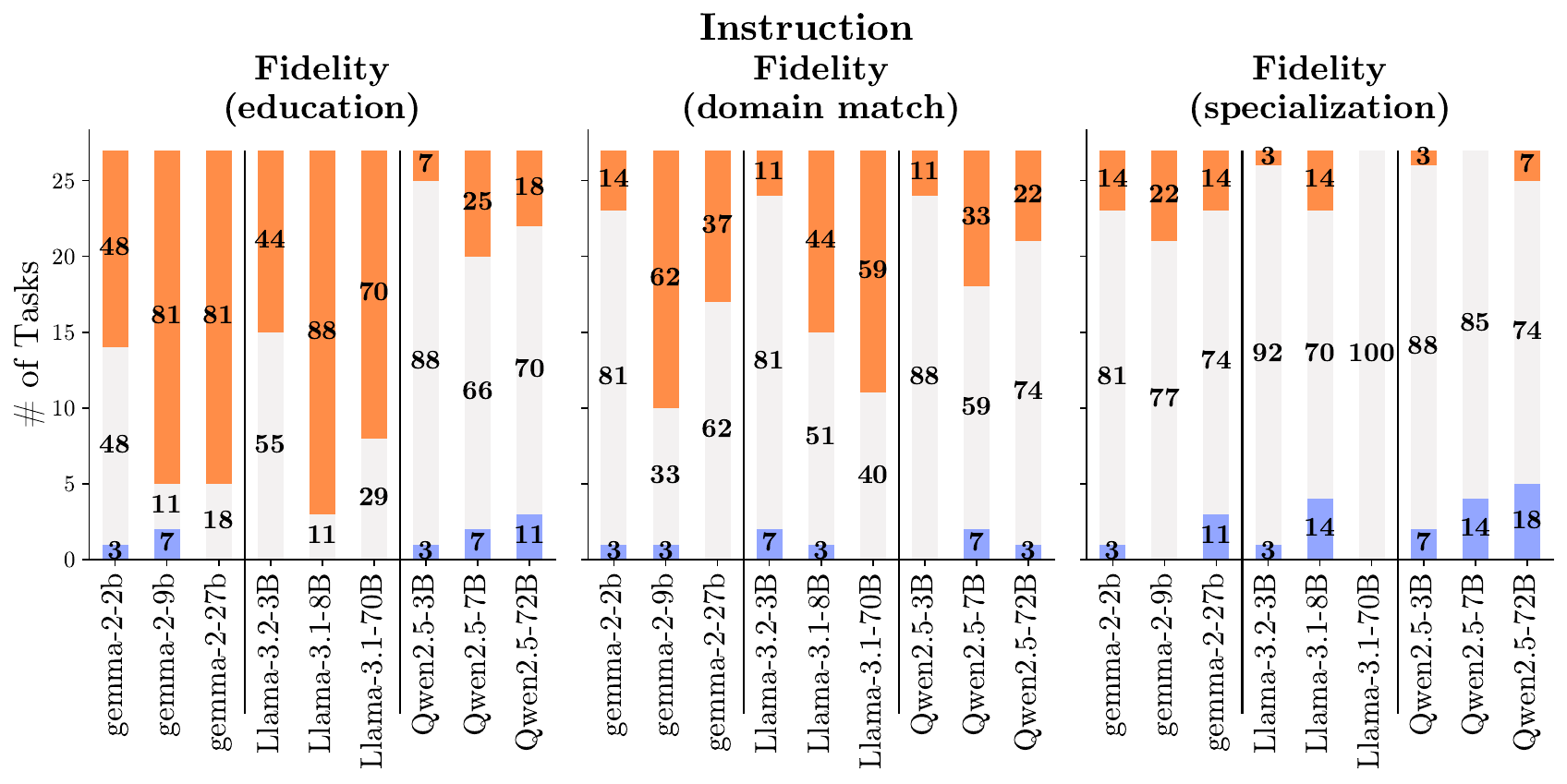}
    \caption{Number of tasks in which the Fidelity metric (with respect to education level, domain match, and expertise specialization) was \textcolor{orange}{positive}, \textcolor{blue}{negative}, or not significant using the Instruction strategy. In-bar annotations indicate the
    percentage of tasks in each category. }
    \label{fig:fid_instruction_agg}
\end{figure}

\begin{figure}[tbhp]
    \includegraphics[width=\linewidth]{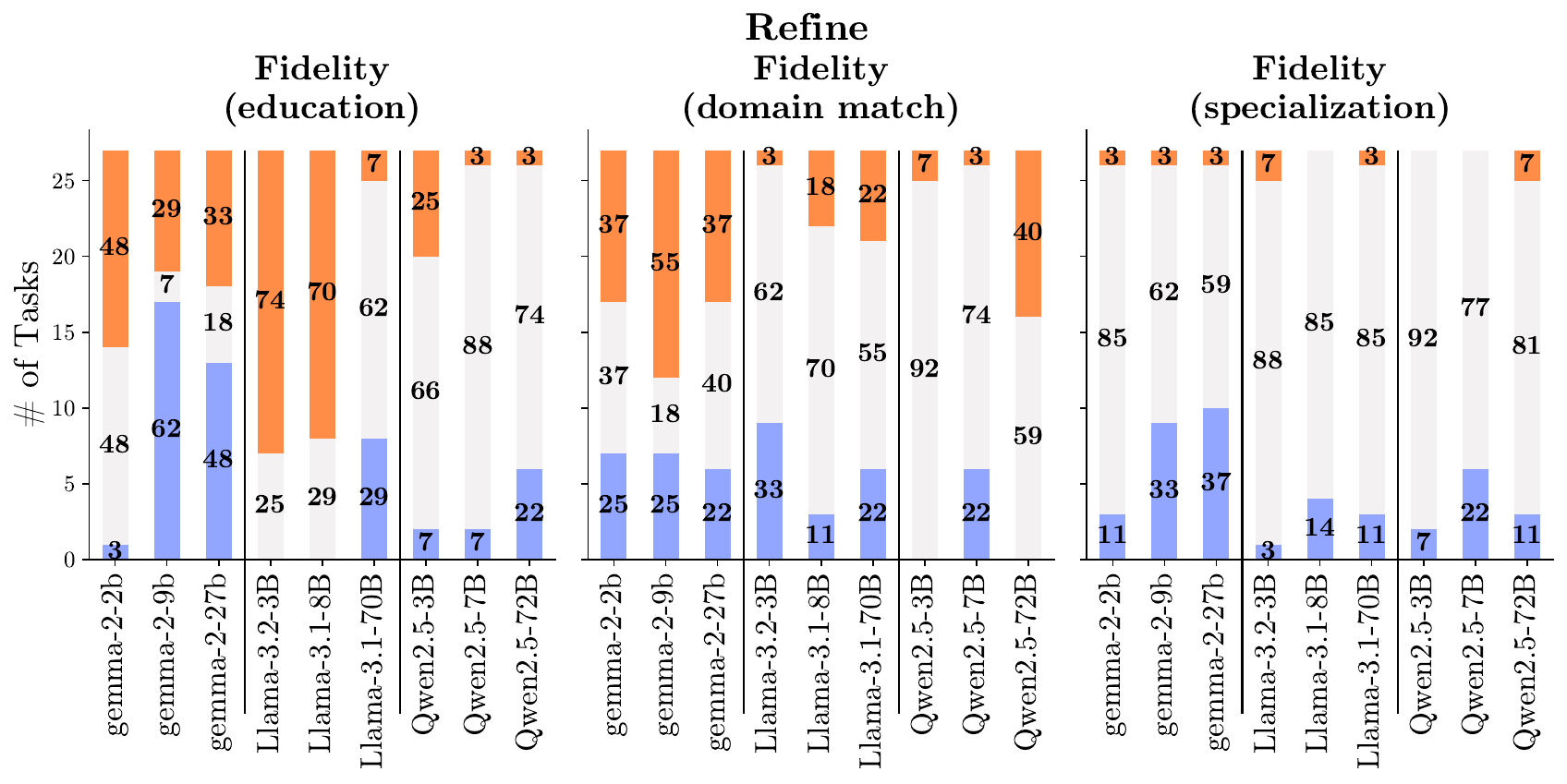}
    \caption{Number of tasks in which the Fidelity metric (with respect to education level, domain match, and expertise specialization) was \textcolor{orange}{positive}, \textcolor{blue}{negative}, or not significant using the Refine strategy. In-bar annotations indicate the
    percentage of tasks in each category. }
    \label{fig:fid_refine_basic_agg}
\end{figure}

\begin{figure}[tbhp]
    \includegraphics[width=\linewidth]{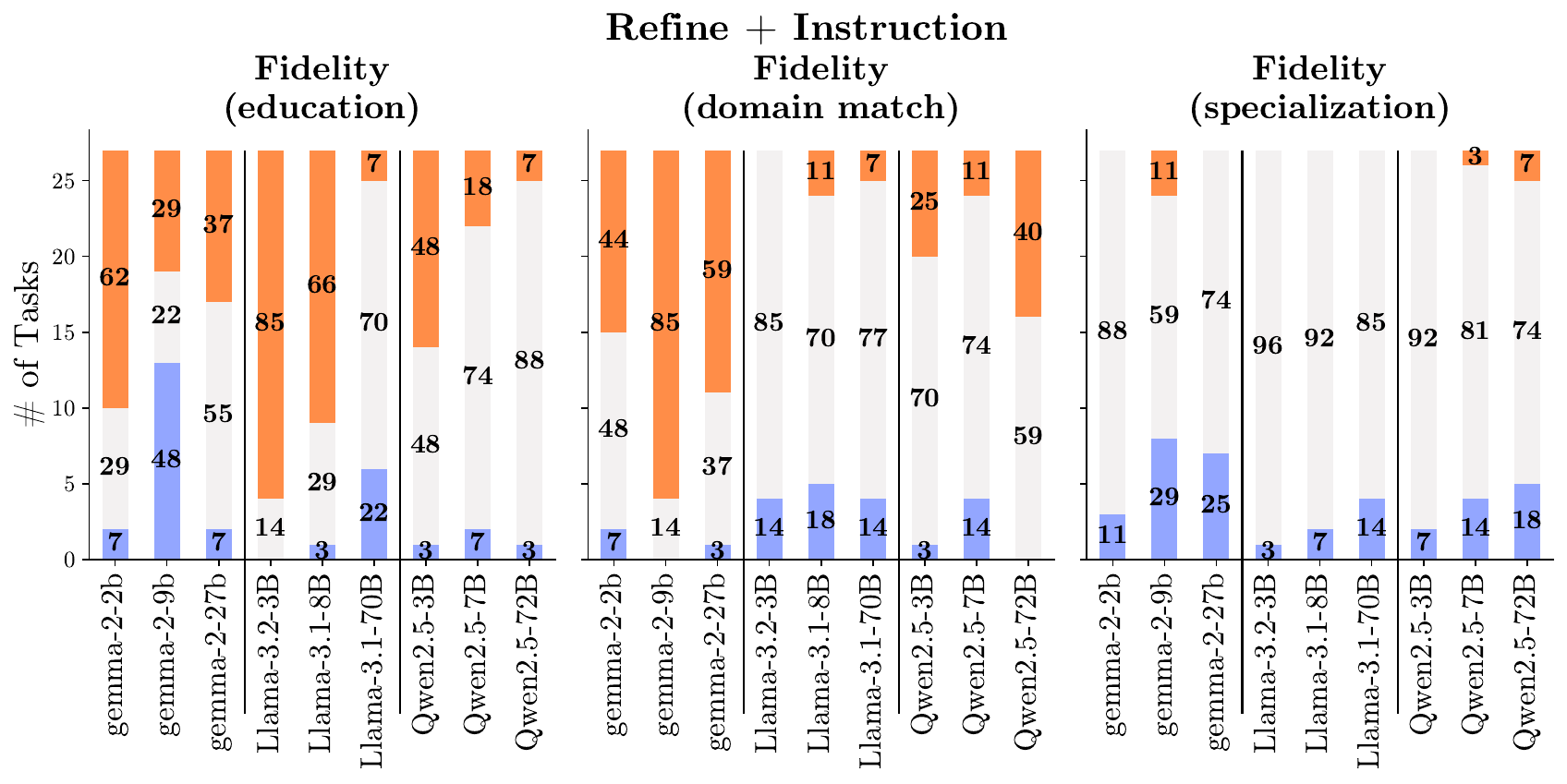}
    \caption{Number of tasks in which the Fidelity metric (with respect to education level, domain match, and expertise specialization) was \textcolor{orange}{positive}, \textcolor{blue}{negative}, or not significant using the Refine + Instruction strategy. In-bar annotations indicate the
    percentage of tasks in each category. }
    \label{fig:fid_refine_agg}
\end{figure}

Figures~\ref{fig:expertise_instruction_agg}-\ref{fig:fid_refine_agg} show aggregate results for each metric and mitigation strategy.

% \subsection{Instruction (refine)}

% \begin{figure*}
%     \includegraphics[width=\linewidth]{media/expertise_aggregate_refine.pdf}
%     \caption{Number of tasks for which the Instruction (refine) prompts with expert personas led to a significant score increase, decrease, or no change, relative to the no-persona baseline. In-bar annotations indicate the percentage of tasks in each category.}
%     \label{fig:expertise_refine_agg}
% \end{figure*}

% \begin{figure}
%     \includegraphics[width=\linewidth]{media/robustness_aggregate_refine.pdf}
%     \caption{Number of tasks for which the Instruction (refine) prompts  with irrelevant personas led to a significant score increase, decrease, or no change, relative to no-persona baseline. In-bar annotations indicate the percentage of tasks in each category.}
%     \label{fig:rob_refine_agg}
% \end{figure}

% \begin{figure}
%     \includegraphics[width=\linewidth]{media/fidelity_aggregate_refine.pdf}
%     \caption{Number of tasks for which the persona performance in Instruction (refine) prompts with respect to education level, domain match, and expertise specialization were (mis)aligned with pre-defined expectations. In-bar annotations indicate the percentage of tasks in each category.}
%     \label{fig:fid_refine_agg}
% \end{figure}

% \subsection{Instruction}

% \begin{figure*}
%     \includegraphics[width=\linewidth]{media/expertise_aggregate_instruction.pdf}
%     \caption{Number of tasks for which the Instruction prompts with expert personas led to a significant score increase, decrease, or no change, relative to the no-persona baseline. In-bar annotations indicate the percentage of tasks in each category.}
%     \label{fig:expertise_instruction_agg}
% \end{figure*}

% \begin{figure}
%     \includegraphics[width=\linewidth]{media/robustness_aggregate_instruction.pdf}
%     \caption{Number of tasks for which the Instruction prompts  with irrelevant personas led to a significant score increase, decrease, or no change, relative to no-persona baseline. In-bar annotations indicate the percentage of tasks in each category.}
%     \label{fig:rob_instruction_agg}
% \end{figure}

% \begin{figure}
%     \includegraphics[width=\linewidth]{media/fidelity_aggregate_instruction.pdf}
%     \caption{Number of tasks for which the persona performance in Instruction prompts with respect to education level, domain match, and expertise specialization were (mis)aligned with pre-defined expectations. In-bar annotations indicate the percentage of tasks in each category.}
%     \label{fig:fid_instruction_agg}
% \end{figure}

% \subsection{Refine

\end{document}